\pdfoutput=1

\documentclass[11pt]{article}

\usepackage[final]{acl}

\usepackage{times}
\usepackage{latexsym}
\usepackage{amsmath, bbm, algorithm, algorithmic}

\usepackage[T1]{fontenc}
\usepackage{subfigure}
\usepackage{graphicx}
\usepackage{amsmath}
\usepackage{amsfonts} 
\usepackage{tabularray}
\usepackage{booktabs}
\usepackage{dsfont}
\usepackage[utf8]{inputenc}

\usepackage{microtype}

\usepackage{inconsolata}

\usepackage{graphicx}

\usepackage{float}
\usepackage{graphicx}
\usepackage{amsmath}
\usepackage{multirow}

\newcommand{\ie}{\textit{i}.\textit{e}.}
\newcommand{\eg}{\textit{e}.\textit{g}.}
\title{Towards Low-Resource Harmful Meme Detection with LMM Agents}

\author{Jianzhao Huang$^{\heartsuit}$\thanks{\; Equal contribution. Hongzhan Lin is the project leader.}, Hongzhan Lin$^{\spadesuit *}$\thanks{\; Corresponding authors.}, Ziyan Liu$^{\heartsuit *}$, Ziyang Luo$^{\spadesuit}$,\\ \textbf{Guang Chen}$^{\heartsuit \dagger}$, \textbf{Jing Ma}$^{\spadesuit \dagger}$ \\
        $^{\heartsuit}$Beijing University of Posts and Telecommunications \\
        $^{\spadesuit}$Hong Kong Baptist University \\
        \texttt{\{cshzlin,majing\}@comp.hkbu.edu.hk},
        \texttt{chenguang@bupt.edu.cn}}

\begin{document}
\maketitle
\begin{abstract}
The proliferation of Internet memes in the age of social media necessitates effective identification of harmful ones. Due to the dynamic nature of memes, existing data-driven models may struggle in low-resource scenarios where only a few labeled examples are available. In this paper, we propose an agency-driven framework for low-resource harmful meme detection, employing both outward and inward analysis with few-shot annotated samples.  Inspired by the powerful capacity of Large Multimodal Models (LMMs) on multimodal reasoning, we first retrieve relative memes with annotations to leverage label information as auxiliary signals for the LMM agent. Then, we elicit knowledge-revising behavior within the LMM agent to derive well-generalized insights into meme harmfulness. By combining these strategies, our approach enables dialectical reasoning over intricate and implicit harm-indicative patterns. Extensive experiments conducted on three meme datasets demonstrate that our proposed approach achieves superior performance than state-of-the-art methods on the low-resource harmful meme detection task.
\end{abstract}

\section{Introduction}
The rise of social media has catalyzed the emergence of a new multimodal entity: the meme. Typically, a meme combines a visual element with concise text, making it highly sharable and capable of quick proliferation across various online platforms. Although often viewed humorously, memes can become vehicles of harm when their mix of image and text is strategically used in the context of political and socio-cultural divisions.

A widely accepted definition of harmful memes\footnote{\color{red}\textbf{Disclaimer:} This paper contains content that may be disturbing to some readers.} is ``multimodal units consisting of an image and embedded text that have the potential to cause harm to an individual, an organization, a community, or society in general''~\cite{sharma2022detecting}. For example, during the COVID-19 pandemic, a frequently shared meme shown in Figure~\ref{fig:meme_1} was created by anti-vaccination groups using a manipulated image of Bill Gates. The widespread dissemination of this multimodal content\footnote{\url{https://www.bbc.com/news/55101238}}, which spread fear about COVID-19 vaccines, significantly harmed Bill Gates' personal reputation and undermined efforts to bolster public immunity. Therefore, it becomes imperative to develop automatic approaches for harmful meme detection to effectively unveil the dark side of memes on social media.

{
\begin{figure}[t]
\subfigure[]{
\begin{minipage}[t]{0.5\linewidth}
\centering
\scalebox{0.75}{\includegraphics[width=4.7cm]{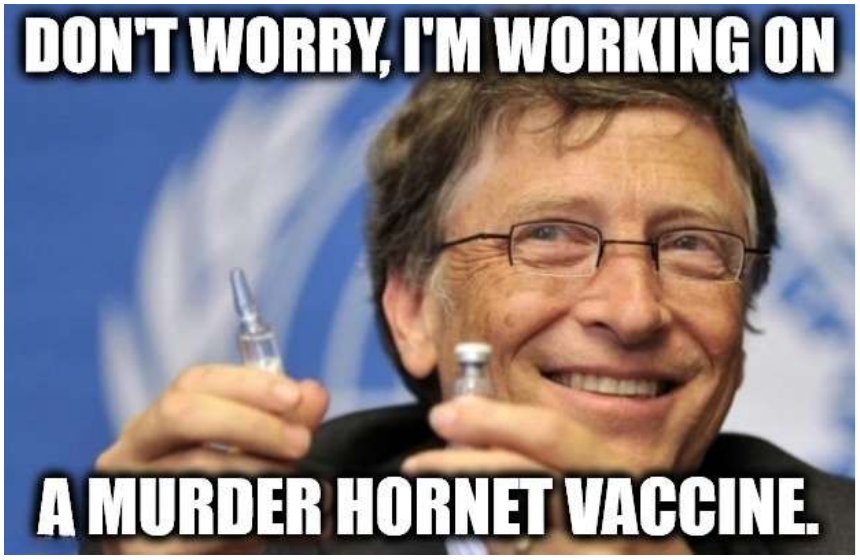}}
\label{fig:meme_1}
\end{minipage}%
}%
\subfigure[]{
\begin{minipage}[t]{0.5\linewidth}
\centering
\scalebox{0.75}{\includegraphics[width=5cm]{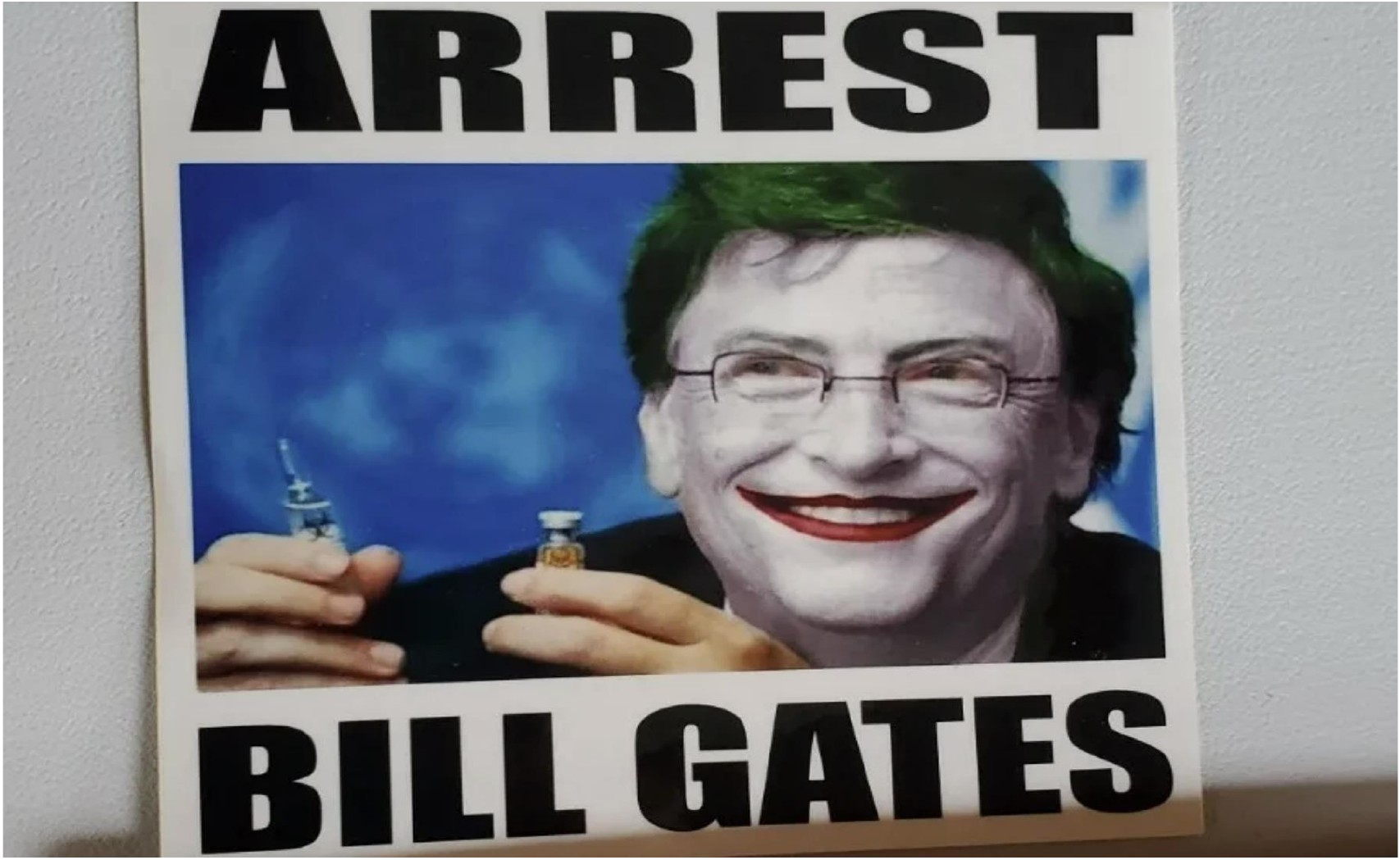}}
\label{fig:meme_2}
\end{minipage}%
}%
\centering
\vspace{-0.4cm}
\caption{Example of trending memes on social media. \textbf{Meme text}: (a) ``\textit{DON'T WORRY, I'M WORKING ON A MURDER HORNET VACCINE.}''; (b) ``\textit{ARREST BILL GATES.}'' }
\label{meme_illustration}
\vspace{-0.4cm}
\end{figure}

Previous studies employing deep neural networks (DNNs) have advanced harmful meme detection~\cite{kiela2020hateful, pramanick2021detecting} by utilizing data-driven multimodal models~\cite{pramanick2021momenta, cao2023prompting, lin2023beneath}, which rely heavily on extensively annotated data for training. However, these DNN-based approaches~\cite{cao2023pro, lin2024explainable} encounter significant challenges when it comes to detecting newly emerging memes related to breaking events, since the rapid development of such harmful memes makes it impractical to swiftly annotate enough training data~\cite{cao2024modularized}. Moreover, the dynamic and evolving nature of Internet memes raises concerns about the sustained effectiveness of traditional data-driven methods for harmful meme detection in real-world scenarios. This highlights the need for more adaptive and responsive approaches in the detection of harmful memes under the low-resource regime where only limited few-shot meme annotations are available.

Intuitively, the key to low-resource harmful meme detection is to strengthen the adaptability to continually changing online memes, and the capacity to respond promptly to new trends and contexts: 1) We posit that the shared characteristics among similar memes could facilitate the adaptability of harmful meme detection, as memes evolve by retaining inherent patterns. For instance, as depicted in Figure~\ref{meme_illustration}, Bill Gates shown in the meme of Figure~\ref{fig:meme_1} could be further incorporated with a joker face, which is extended to create a new harmful variant of Figure~\ref{fig:meme_2}. 2) On the other hand, rather than using gradient descent to update model parameters, we resort to imitating human learning processes that accumulate experience without modifying the model's weights, thus mitigating the risk of overfitting to sparse annotations of harmful memes. As an example, to illustrate the harmfulness of the memes in Figure~\ref{meme_illustration}, a human checker needs the reasoning knowledge to gather the experience that Bill Gates is frequently vilified in harmful memes by anti-vaccination campaigners due to his advocacy for vaccine development. Thus we devise a gradient-free approach to capture common features of harmful memes and derive insights from limited annotated training meme data for better generalization in a low-resource learning context.

Inspired by the powerful capacity of LMMs for reasoning with contextual background knowledge~\cite{brown2020language, liu2023visual}, we propose a novel agentic approach: \textbf{\textsc{LoReHM}}, towards \underline{\textbf{Lo}}w-\underline{\textbf{Re}}source \underline{\textbf{H}}armful \underline{\textbf{M}}eme detection by regarding LMMs as agents. To this end, we propose capturing the harmfulness of memes by employing both outward and inward analysis with limited few-shot annotated meme samples. Specifically, 1) for looking outward at a meme, we first retrieve its similar memes with labels to leverage the annotation information as explicit auxiliary signals for harmfulness preference, thereby facilitating the decision-making of the LMM agent. 2) In terms of looking inward at a meme, we employ the fundamental skill of human learning, positioning the LMM as a learner agent to derive well-generalized insights from its failed attempts on the limited annotated meme samples, which aims to capture the implicit harmfulness meanings not conveyed through the superficial texts and images of memes. 3) Finally, we combine the strategies of looking outward and inward to enable the LMM agent to perform the final harmfulness inference. In this manner, we enhance the LMM's ability as a trustworthy agent to detect harmful content concealed in the intrinsic multimodal information in memes. Our contributions are summarized as follows in three folds:
\begin{itemize}
    \item To our best knowledge, we are the first to alleviate the low-resource issue of harmful meme detection from a fresh gradient-free perspective on harnessing advanced LMMs.\footnote{Our source code is available at \url{https://github.com/Jianzhao-Huang/LoReHM}.}
    \vspace{-0.21cm}
    \item We propose a novel agency-driven approach for low-resource harmful meme detection, to augment the LMM agent with harm-indicated signals from retrieval of outward relatively similar memes and insight of inward multimodal knowledge-revising, which facilitates harmfulness inference in a few-shot regime.
    \vspace{-0.21cm}
    \item Extensive experiments conducted on three meme datasets confirm that our agentic paradigm could yield superior few-shot performance than previous state-of-the-art baselines for low-resource harmful meme detection.
\end{itemize}

\section{Related Work}
\subsection{Harmful Meme Detection}
Harmful meme detection is an expanding field, bolstered by large meme benchmarks~\cite{kiela2019supervised, pramanick2021detecting, lin2024goat}, and initiatives like the Hateful Memes Challenge~\cite{kiela2020hateful} by Facebook for detecting memes in hate speech~\cite{das2020detecting}. These developments have spurred research into detecting harmful memes~\cite{pramanick2021detecting}, a task complicated by their multimodal nature, which often involves both texts and images. As unimodal methods like BERT~\cite{ devlin2018bert} or Faster R-CNN~\cite{ren2016faster} fall short in addressing these complexities, recent studies have increasingly turned to multimodal approaches to improve the detection performance of harmful memes.

Previous studies have employed classical two-stream models that integrate textual and visual features, which are learned from text and image encoders, typically utilizing attention-based mechanisms and multimodal fusion techniques for classifying harmful memes~\cite{kiela2019supervised, kiela2020hateful, suryawanshi2020multimodal, pramanick2021momenta}. Another branch was fine-tuning pre-trained multimodal models specifically for the task~\cite{lippe2020multimodal, muennighoff2020vilio, velioglu2020detecting, hee2022explaining}. Recent efforts have also sought to explore the use of data augmentation techniques~\cite{zhou2021multimodal, zhu2022multimodal}, ensemble methods~\cite{zhu2020enhance, velioglu2020detecting, sandulescu2020detecting}, harmful target disentanglement~\cite{lee2021disentangling}, and prompt-based tuning~\cite{cao2023prompting, ji2023identifying, cao2023pro}.
Lately, \citet{lin2023beneath} proposed to distill multimodal reasoning knowledge from Large Language Models (LLMs) to detect harmful memes.

However, such data-driven approaches fail to detect harmful evolving memes in low-resource regimes because they often require sizeable training data unavailable for emerging events. Although a recent work~\cite{cao2024modularized} employed low-rank adaptation (LoRA)~\cite{hu2021lora} for detecting harmful memes in the few-shot setting, we delve into the low-resource harmful meme detection without updating any model weights, which utilizes multimodal LLMs (\ie, LMMs) in an agency-driven manner by leveraging both LMM’s internal knowledge and multimodal retrieval-augmented generation to examine the harmfulness of a meme.

\subsection{LLM Agent}
Retrieval-augmented generation enriches the input space of LLMs with retrieved text passages~\citep{guu2020retrieval, lewis2020retrieval}, resulting in significant enhancements in knowledge-intensive tasks and decision-making agents, either through fine-tuning or utilization with off-the-shelf LLMs~\citep{liu2022makes, schick2023toolformer, ram2023context, izacard2023atlas, jiang2023active, asai2023self, gao2023enabling, wang2023self}. 
The integration of LLMs as agents spans various domains, including code generation and game-playing, showcasing their robust planning and reasoning capabilities in diverse settings~\cite{wang2023voyager, yao2022react, shen2023hugginggpt, mu2023embodiedgpt, hong2023metagpt, liu2023agentbench, zhao2024expel, sun2023adaplanner, qian2023communicative}. These advancements underscore the ability of LLMs to tackle complex tasks with minimal supervision. Concurrently, self-improvement methodologies~\cite{chen2022codet, chen2023teaching, shinn2024reflexion, madaan2023self} have emerged, leveraging feedback-driven processes to iteratively refine generated outputs. In contrast to many previous agent tasks that operate within environments providing real feedback, harmful meme detection lacks an environment that supplies the agent with authentic responses. In this work, we focus on a novel agentic paradigm by devising the proprietary integration of the vision-language retrieval-augmented and self-improvement mechanisms for LMMs~\cite{liu2023visual, OpenAI2023GPT4TR}, to detect harmful memes with limited few-shot annotations, a realistic yet urgent task that is inherently a binary multimodal classification challenge.

\section{Our Approach}
\begin{figure*}
    \centering
    \includegraphics[width=1\linewidth]{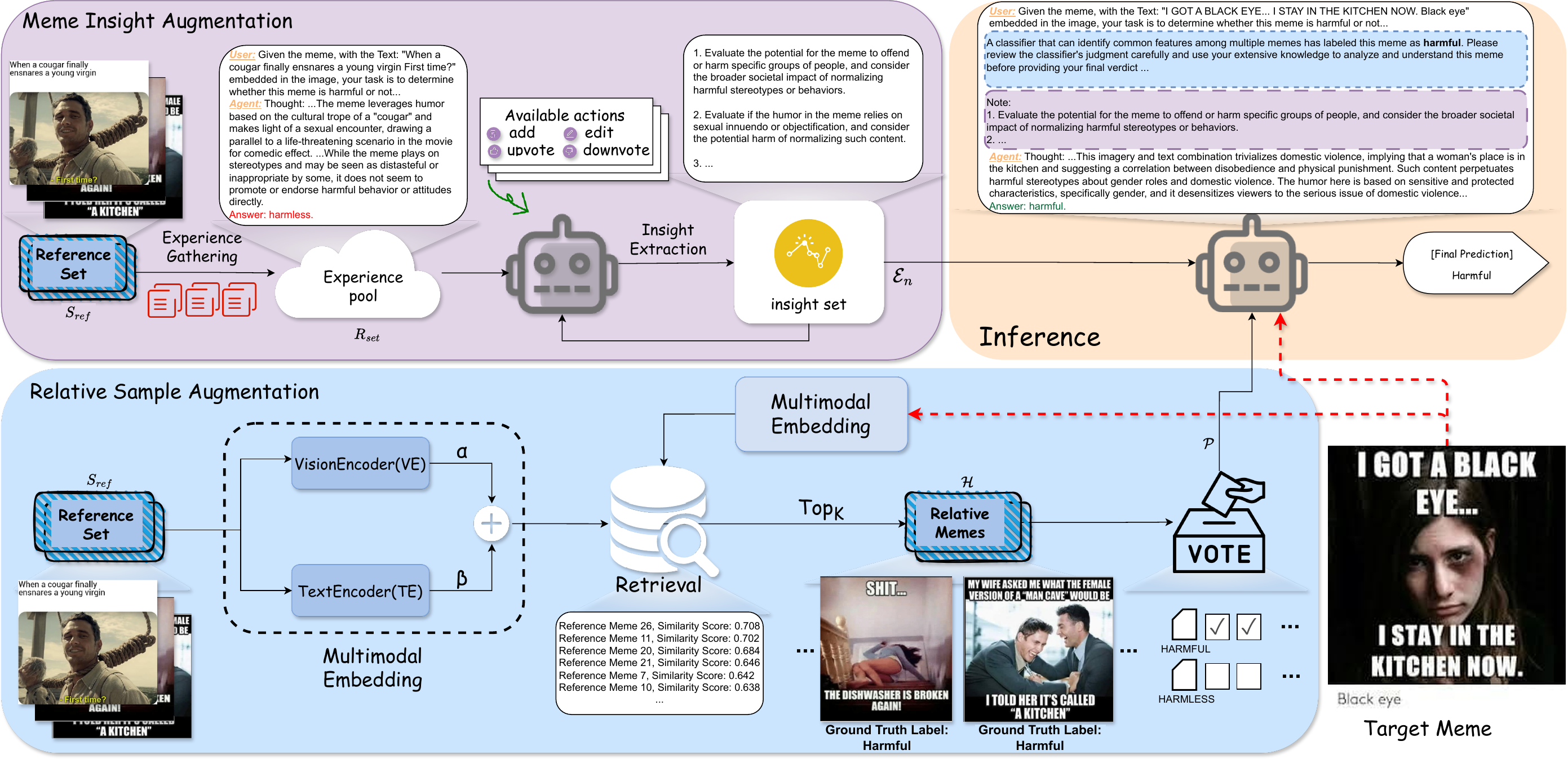}
    \vspace{-0.7cm}
    \caption{An overview of our framework, \textsc{LoReHM}, for low-resource harmful meme detection.}
    \label{fig:main_figure}
    \vspace{-0.3cm}
\end{figure*}
\subsection{Overview}
\paragraph{Problem Statement.} We define a harmful meme detection dataset as a set of memes where each meme $M=\{\mathcal{I},\mathcal{T},y\}$ is a triplet representing an image $\mathcal{I}$ that is associated with a text $\mathcal{T}$, and a ground-truth harmfulness label $y \in \{\texttt{harmful}, \texttt{harmless}\}$. In this work, to investigate low-resource harmful meme detection with LMM agents, we convert the harmful meme detection task into a natural language generation paradigm, where our model takes the image $\mathcal{I}$ and text $\mathcal{T}$ as the input and generates a text sequence to clearly express whether the meme is harmful or not.

The scarcity of high-quality labeled data is gradually becoming the norm with the rapid merging and evolution of memes~\cite{sharma2022detecting}. In this work, we define a low-resource setting where we have a test set $S_{\text{test}}$, and a very small training set $S_{\text{train}}$ comprising $N$ samples. Since our method is gradient-free, we denote $S_{\text{train}}$ as the reference set $S_{\text{ref}}$. Our objective is to optimize the model based on the very limited reference set $S_{\text{ref}}$, enabling it to effectively generalize to the test dataset $S_{\text{test}}$.

Our core idea is to regard the LMM as an agent, and then enhance the agent's memory by using information from relative memes and insights derived from the agent's past unsuccessful evaluations of memes, thereby improving its performance on harmful meme detection in low-resource scenes. We enable the LMM agent to fully capitalize on a very limited labeled dataset through two strategies: 1) Relative Sample Augmentation (\S\ref{RSA}) for learning by looking outward, and 2) Meme Insight Augmentation (\S\ref{MIA}) for learning by looking inward. Both strategies allow for the efficient extraction of harmfulness cues from limited annotated memes. This empowers the LMM agent to perform more precisely and effectively in the task of harmful meme detection under low-resource regimes. The overview of our framework is shown in Figure~\ref{fig:main_figure}.

\subsection{Relative Sample Augmentation}
\label{RSA}

With the aid of LMMs, it becomes plausible to offer a powerful few-shot performance of low-resource harmful meme detection by In-context Learning (ICL)~\cite{brown2020language}. However, the ICL approach involves integrating few-shot demonstrations with test instances repeatedly, which incurs substantial computational overhead at each inference step. This can be impractically burdensome for real-world applications. 
Generally, Internet memes evolve dynamically yet exhibit inherent patterns~\cite{baran2001prologue}. In this section, drawing inspiration from retrieval-augmented generation, we propose to capitalize on the highly shared characteristics of harmful memes to enhance the detection performance of LMMs in the low-resource setting. Specifically, our approach utilizes explicit labels of top-retrieved meme samples, derived from multimodal information retrieval, as auxiliary signals.
\subsubsection{Relative Sample Retrieval}
For a meme sample $M$, we first generate visual textual embeddings, and then fuse them with a fixed ratio to produce a multimodal representation:
{
\begin{equation}
\label{encoding}
Emb=\alpha \cdot \operatorname{VE}(\mathcal{I})+\beta \cdot \operatorname{TE}(\mathcal{T}),
\end{equation}}
where $Emb$ is the multimodal embedding of $M$, $\operatorname{VE}(\cdot)$ and $\operatorname{TE}(\cdot)$ represent the frozen pre-trained vision and text Transformer encoders, respectively, while $\alpha$ and $\beta$ denote the ratio factors. Then, we can encode all the meme samples from both the reference set $S_\text{ref}$ and the test set $S_\text{test}$ as Equation~\ref{encoding}, to obtain the fixed embeddings of meme samples.

Given a target meme $M_\text{test}$ in the test set $S_\text{test}$, to retrieve the most relative candidate meme samples in the reference set $S_\text{ref}$ to the target meme, we calculate the similarity between the embedding of the target meme and the embedding of each meme $M_{\text{ref}}$ in $S_\text{ref}$ as follows:
{
\begin{equation}
\label{calculate_similarity}
d = \text{sim}(M_{\text{ref}}, M_{\text{test}}),
\end{equation}}
where \(d\) represents the normalized cosine similarity score between the meme \(M_{\text{ref}}\) in the reference set \(S_\text{ref}\) and the target meme \(M_{\text{test}}\). The function \(\text{sim}(\cdot)\) computes the similarity by comparing the multimodal embeddings of a pair of memes. Thus all the similarity scores of the candidate memes in the reference set to the target meme could form the similarity vector $D = \{d \mid M_{\text{ref}} \in S_\text{ref}\} \in \mathbb{R}^N$.

Afterwards, we select the top \(K\) memes\footnote{Note that \(K\) should be an odd number ($K < N$).} from $S_\text{ref}$ with the highest similarity scores as the relative memes to the target meme $M_\text{test}$:
{
\begin{equation}
\label{retrieve topk}
    \mathcal{H} = \{M_{\text{ref}} \mid d \in Top_K(D) \},
\end{equation}}
where \(\mathcal{H}\) is the set of the \(K\) memes from $S_\text{ref}$ most similar to the target meme $M_\text{test}$. The function $Top_K(\cdot)$ ranks and selects the top \(K\) highest scores from the set \(D\) of similarity scores, thereby identifying the 
\(K\) most relevant memes in the reference set $S_\text{ref}$ to the target meme $M_\text{test}$.

\subsubsection{Voting Mechanism}
The memes in the retrieved set $\mathcal{H}$ exhibit a high degree of multimodal similarity with the target meme. This similarity is a crucial indicator of the common salient multimodal features shared between the memes, which are often products of meme evolution and integration. Therefore, we exploit the harmfulness of the retrieved set to explicitly infer the predicted harmfulness label of the target meme. 

Based on the retrieved set $\mathcal{H}$ where the $K$ memes are well-annotated, we employ a voting mechanism to obtain a preliminary prediction for the target meme according to the polarity of the ground-truth labels of memes in the retrieved set:
{
\begin{equation}
\label{voting}
\small
\mathcal{P} = 
\begin{cases} 
\texttt{harmful} & \text{if } \sum_{i=1}^K \mathds{1}(\mathcal{H}_i^y = \texttt{harmful}) > \frac{K}{2} \\
\texttt{harmless} & \text{otherwise}
\end{cases},
\end{equation}}
where $\mathcal{P}$ is the preliminary prediction of the target meme, $\mathcal{H}_i^y$ means the ground truth label of the $i$-th meme in $\mathcal{H}$. In this way, the explicit harmfulness labels of the annotated memes in the retrieved set could be leveraged as the prior harm-indicated signals for the low-resource harmful meme detection.

\subsection{Meme Insight Augmentation}
\label{MIA}
Although Relative Sample Augmentation as \S\ref{RSA} can enhance the LMM agent's ability by providing explicit harmfulness references as prior for the target meme, previous literature has shown that memes generally contain implicit harmful information that is more subtle and challenging for the LMM to identify~\cite{lin2023beneath}. To address this, we propose enabling the LMM agent to better capture implicit harmfulness in memes by looking inward through multimodal knowledge revising.
\subsubsection{Experience Gathering}
To gather the experience that serves as the basis for insight extraction, we first evaluate the harmfulness of each $M_\text{ref}$ in the reference set $S_\text{ref}$ in a zero-shot manner. For performance and explainability, we choose Chain-of-Thought (CoT) as the base planning algorithm~\cite{kojima2022large} and feed the image $\mathcal{I}_\text{ref}$ and text $\mathcal{T}_\text{ref}$ of $M_\text{ref}$ to the agent:
{
\begin{equation}
traj = \{ \operatorname{LMM}(X_{\text{CoT}}, \mathcal{I}_\text{ref}, \mathcal{T}_\text{ref}) \mid M_\text{ref} \in S_{\text{ref}} \},
\end{equation}}
where $traj$ is the trajectories gathered through the LMM agent's direct judgment on the meme $M_\text{ref}$ in $S_\text{ref}$, $\operatorname{LMM}(\cdot)$ denotes the LMM agent, $X_\text{CoT}$ is the zero-shot CoT instruction designed as:

``\textit{Given the meme, with the Text: \{$\mathcal{T}_\text{ref}$\} embedded in the image \{$\mathcal{I}_\text{ref}$\}, your task is to determine whether this meme is harmful or not, in order to maintain the harmlessness and integrity of information on the Internet. Please leverage your extensive knowledge to deeply analyze and understand this meme, and give your final judgment. Your output should strictly follow the format: \{Thought: [Your analysis] Answer: [harmful/harmless]\}.}''

Furthermore, in order to make the LMM agent's self-reflection more accurate and efficient, we obtain a self-reflect set that contains only erroneous trajectories by comparing the answer indicated by the trajectory ${traj}$ and the ground truth label $y_\text{ref}$ corresponding to $M_{\text{ref}}$ as follows:
{
\begin{equation}
R_{\text{set}} = \{ traj \mid \overset{\longrightarrow}{traj} \neq y_\text{ref}, \, M_\text{ref} \in S_{\text{ref}} \},
\end{equation}}
where $R_{\text{set}}$ is a self-reflect set that contains $n$ erroneous trajectories ($n < N$), 
$\overset{\longrightarrow}{traj}$ is the answer indicated by the trajectory $traj$. The self-reflect set $R_{\text{set}}$ could be regarded as an experience pool to prioritize challenging meme examples in the reference set $S_\text{ref}$, which were misjudged by the zero-shot inference of the LMM agent, while avoiding excessive attention to trivial examples that have already been correctly detected by the LMM agent.

\subsubsection{Insight Extraction}
To derive general insights about the harmfulness meaning of memes, we first initialize an empty set of insights, which is denoted as \( \mathcal{E}_0 \), then iteratively feed the failed trajectories from the experience pool (\ie, the self-reflect set $R_{\text{set}}$) into the LMM, prompting the LMM to reflect on these trajectories.

\begin{table*}[!t]
    \centering
\resizebox{0.9\textwidth}{!}{\begin{tabular}{@{}l||cc|cc|cc@{}}
\toprule
Dataset         & \multicolumn{2}{c|}{HarM}                  & \multicolumn{2}{c|}{FHM}                        & \multicolumn{2}{c}{MAMI}                     \\ \midrule
Model           & Accuracy                 & Macro-$\emph{F}_1$                & Accuracy                 & Macro-$\emph{F}_1$                     & Accuracy                 & Macro-$\emph{F}_1$               \\ \midrule \midrule
{PromptHate}~\cite{cao2023prompting}       & 63.56               & 61.94                 & 54.80                & \multicolumn{1}{c|}{54.78} &  56.84           & 55.79                     \\
{\textsc{Mr.Harm}}~\cite{lin2023beneath}    &  71.56               & 70.62                 & 55.00               & \multicolumn{1}{c|}{51.79 } & 57.20                & 56.25                \\ 
{Pro-Cap}~\cite{cao2023pro}     & 71.47                & 69.45                 & 56.60                & \multicolumn{1}{c|}{56.14} & 62.31                & 61.48                     \\ \midrule
{OPT-30B}~\cite{zhang2022opt} & 66.95                & 64.72                  & 54.20                & \multicolumn{1}{c|}{50.82} & 63.40                 & 63.40                 \\
{OpenFlamingo-9B}~\cite{awadalla2023openflamingo}      & 66.95                & 59.36                 &  51.60               & \multicolumn{1}{c|}{51.52} & 52.70                & 46.80                     \\
{Mod-HATE}~\cite{cao2024modularized}            &                 71.19&                  69.64&                  57.60& \multicolumn{1}{c|}{53.88} &                 69.05&                      68.78\\ \midrule
LLaVA-34B~\cite{liu2024llava}         &                 67.80& 62.60& 63.80& \multicolumn{1}{c|}{63.74} &                 74.60&                                74.52\\
GPT-4o~\cite{OpenAI2023GPT4TR}         & 71.75                & 70.23                 & \underline{66.60}                & \multicolumn{1}{c|}{\underline{65.74}} & \underline{80.80}                & \underline{80.52}                                \\ \midrule 
\textsc{LoReHM} (LLaVA-34B)          & \multicolumn{1}{c}{\underline{73.73}} & \multicolumn{1}{c|}{\underline{70.86}} & \multicolumn{1}{c}{\underline{}65.60} & \multicolumn{1}{c|}{\textbf{}65.59} & \multicolumn{1}{c}{\textbf{}75.40} & \textbf{}75.28 \\
\textsc{LoReHM} (GPT-4o)          & \multicolumn{1}{c}{\textbf{74.57}} & \multicolumn{1}{c|}{\textbf{72.98}} & \multicolumn{1}{c}{\textbf{70.20}} & \multicolumn{1}{c|}{\textbf{70.14}} & \multicolumn{1}{c}{\textbf{83.00}} & \textbf{82.98}\\\bottomrule
\end{tabular}}
\vspace{-0.2cm}
    \caption{Low-resource harmful meme detection results on three datasets. The accuracy and macro-averaged F1 scores (\%) are reported as the metrics. The best and second test results are in bold and underlined, respectively.}
    \label{tab:main_results}
    \vspace{-0.4cm}
\end{table*}

Specifically, for gaining general insights into low-resource harmful meme detection, rather than obsessed with a specific failed meme, we prompt the LMM to perform a series of operations (\textit{ADD}, \textit{DOWNVOTE}, \textit{UPVOTE}, \textit{EDIT}) on the insight set. This approach allows for a broader understanding instead of directly outputting simple insights based on the erroneous trajectory of a particular meme in $R_{\text{set}}$.
In each iteration, the LMM takes a trajectory from the experience pool, then analyzes the reasons for its failure, and in conjunction with the current insight set, determines the operations to be performed on the current insight set as follows:
{
\begin{equation}
\mathcal{O}_i = \operatorname{LMM}(X_\text{Reflect}, traj_i, \mathcal{E}_{i-1}),
\end{equation}}
where $traj_i$ is the $i$-th ($1 \le i \le n$) trajectory in $R_{\text{set}}$, $\mathcal{E}_{i-1}$ is the current insight set in the $i$-th iteration, and $\mathcal{O}_i$ denotes the operations produced in the $i$-th iteration.
$X_\text{Reflect}$ is the reflection instruction designed as detailed in the Appendix \S\ref{sec:ID}.

Subsequently, these operations are applied to the current insight set $\mathcal{E}_{i-1}$, thereby updating it as:
{
\begin{equation}
\mathcal{E}_{i} = \mathcal{O}_i(\mathcal{E}_{i-1}).
\end{equation}}
The performed operations include: \textit{ADD}, to introduce a new generic insight; \textit{DOWNVOTE}, to downvote an existing insight; \textit{UPVOTE}, to agree with an existing insight; and \textit{EDIT}, to modify the contents of an existing insight. An added insight will have an initial importance count, which will increment if subsequent operations \textit{UPVOTE} or \textit{EDIT} are applied to it, and decrement if \textit{DOWNVOTE}. We denote $\mathcal{E}_n$ as the final insight set.

\subsection{Inference}
For the given target meme $M_\text{test}$, we can attain both the preliminary prediction $\mathcal{P}$ in \S\ref{RSA} and the insight set $\mathcal{E}_n$ in \S\ref{MIA}. Finally, the agent utilizes the preliminary assessment \textbf{$\mathcal{P}$} as prior, under the guidance of the insight set $\mathcal{E}_n$, to evaluate whether the meme is harmful or not, culminating in a final judgment output as $\operatorname{LMM}(X_{\text{CoT}}, \mathcal{I}_\text{test}, \mathcal{T}_\text{test}, \mathcal{P}, \mathcal{E}_n)$. We set the number $N$ of memes in the reference set $S_\text{ref}$ as 50, and the number $K$ of memes in the retrieved set $\mathcal{H}$ as 5. We select LLaVA-34B~\cite{liu2023visual} and GPT-4o~\cite{OpenAI2023GPT4TR} as the two representative backbones of the LMM agent.

\section{Experiments}
\subsection{Experimental Setup}
\paragraph{Datasets} We use three publicly available meme datasets for evaluation: (1) HarM~\cite{pramanick2021detecting}, (2) FHM~\cite{kiela2020hateful}, and (3) MAMI~\cite{fersini2022semeval}. HarM consists of memes related to COVID-19. FHM was released by Facebook as part of a challenge to crowd-source multimodal harmful meme detection in hate speech solutions. MAMI encompasses a dataset of memes that are predominantly derogatory towards women, exemplifying typical subjects of online vitriol. Different from FHM and MAMI, where each meme was labeled as \textit{harmful} or \textit{harmless}, HarM was originally labeled with three classes: \textit{very harmful}, \textit{partially harmful}, and \textit{harmless}. For a fair comparison, we merge the \textit{very harmful} and \textit{partially harmful} memes into the \textit{harmful} class, following the setting of recent work~\cite{pramanick2021momenta, cao2023prompting, lin2023beneath}.
\paragraph{Baselines} We compare \textsc{LoReHM} with several state-of-the-art (SoTA) systems for low-resource harmful meme detection: 1) \textsf{PromptHate}~\cite{cao2023prompting}; 2) \textsf{\textsc{Mr.Harm}}~\cite{lin2023beneath};
3) \textsf{Pro-Cap}~\cite{cao2023pro}; 4) \textsf{OPT-30B}~\cite{zhang2022opt}; 5) \textsf{OpenFlamingo-9B}~\cite{awadalla2023openflamingo}; 6) \textsf{Mod-HATE}~\cite{cao2024modularized}; 7) \textsf{LLaVA-34B}~\cite{liu2023visual}; 8) \textsf{GPT-4o}~\cite{OpenAI2023GPT4TR}; 9) \textsf{\textsc{LoReHM} (*)}: Our proposed agentic approach based on LLaVA-34B and GPT-4o. We use the accuracy and macro-averaged F1 (dominant) scores as the evaluation metrics.

The data statistics, baseline descriptions and model implementation are detailed in the Appendix \S\ref{sec:datasets}, \S\ref{sec:baselines}, and \S\ref{sec:ID}, respectively.

\begin{table*}[t]
    \centering
\resizebox{0.95\textwidth}{!}{\begin{tabular}{@{}ll||cc|cc|cc@{}}
\toprule
Dataset &        & \multicolumn{2}{c|}{HarM}                  & \multicolumn{2}{c|}{FHM}                        & \multicolumn{2}{c}{MAMI}                     \\ \midrule
Model &            & Accuracy                 & Macro-$\emph{F}_1$                & Accuracy                 & Macro-$\emph{F}_1$                     & Accuracy                 & Macro-$\emph{F}_1$               \\ \midrule \midrule
\multirow{5}{*}{LLaVA-34B} & w/ 0-shot Prompt        & \multicolumn{1}{c}{65.82} & \multicolumn{1}{c|}{60.02} & \multicolumn{1}{c}{64.00} & \multicolumn{1}{c|}{63.51} & \multicolumn{1}{c}{72.20} & {72.16} \\
& w/ 50-shot ICL          & \multicolumn{1}{c}{67.80} & \multicolumn{1}{c|}{62.60 } & \multicolumn{1}{c}{63.80} & \multicolumn{1}{c|}{63.74 } & \multicolumn{1}{c}{74.60} & {74.52} \\
& w/ Relative Sample Augmentation & \multicolumn{1}{c}{74.58} & \multicolumn{1}{c|}{70.02} & \multicolumn{1}{c}{60.00} & \multicolumn{1}{c|}{59.98} & \multicolumn{1}{c}{71.60} & {71.45} \\
& w/ Meme Insight Augmentation & \multicolumn{1}{c}{68.93} & \multicolumn{1}{c|}{65.04} & \multicolumn{1}{c}{64.80} & \multicolumn{1}{c|}{64.18} & \multicolumn{1}{c}{73.60} & {73.54} \\
& w/ \textsc{LoReHM} & \multicolumn{1}{c}{73.73} & \multicolumn{1}{c|}{70.86} & \multicolumn{1}{c}{65.60} & \multicolumn{1}{c|}{65.59} & \multicolumn{1}{c}{75.40} & {75.28} \\ \hline
\multirow{5}{*}{GPT-4o} & w/ 0-shot Prompt         & \multicolumn{1}{c}{67.23} & \multicolumn{1}{c|}{63.29} & \multicolumn{1}{c}{65.00} & \multicolumn{1}{c|}{63.19} & \multicolumn{1}{c}{80.50} & {80.34} \\
& w/ 50-shot ICL         & \multicolumn{1}{c}{71.75} & \multicolumn{1}{c|}{70.23} & \multicolumn{1}{c}{66.60} & \multicolumn{1}{c|}{65.74} & \multicolumn{1}{c}{80.80} & {80.52} \\
& w/ Relative Sample Augmentation & \multicolumn{1}{c}{72.03} & \multicolumn{1}{c|}{70.32} & \multicolumn{1}{c}{67.20} & \multicolumn{1}{c|}{66.58} & \multicolumn{1}{c}{81.60} & {81.44} \\
& w/ Meme Insight Augmentation & \multicolumn{1}{c}{70.90} & \multicolumn{1}{c|}{69.12} & \multicolumn{1}{c}{67.80} & \multicolumn{1}{c|}{67.70} & \multicolumn{1}{c}{80.60} & {80.49} \\
& w/ \textsc{LoReHM} & \multicolumn{1}{c}{74.57} & \multicolumn{1}{c|}{72.98} & \multicolumn{1}{c}{70.20} & \multicolumn{1}{c|}{70.14} & \multicolumn{1}{c}{83.00} & {82.98} \\
\bottomrule
\end{tabular}}
\vspace{-0.2cm}
    \caption{Ablation studies on our proposed framework based on different LMM agents.}
    \label{tab:ablation}
    \vspace{-0.4cm}
\end{table*}

\subsection{Harmful Meme Detection Performance}
Table~\ref{tab:main_results} illustrates the performance of our proposed method \textsc{LoReHM} versus all the compared baselines for low-resource harmful meme detection. It is observed that: 1) The performance of the baselines in the first group is relatively poor due to their reliance on fully data-driven paradigms. To ensure fair comparisons in the low-resource few-shot regime, all the baselines are trained using the same amount of limited annotated meme data. 2) For the second group, the baselines are low-resource and LMM-based. Both OPT-30B and OpenFlamingo-9B are LMMs with pre-training, while Mod-HATE is based on the LoRA tuning specific to the task. We can find that neither general pre-training nor specific LoRA tuning could enhance the performance of low-resource harmful meme detection in the few-shot setting. 3) In terms of the two of the most powerful cutting-edge LMMs in the third group, LLaVA-34B is the representative open-source LMM with instruction tuning while GPT-4o is closed-source and enhanced by reinforcement learning with human feedback~\cite{ouyang2022training}. Compared with the baselines in the second group, the two SoTA LMMs by few-shot ICL prompts show performance improvement in general, since their advanced training strategies optimize alignment with human values and better adapt to in-context learning. Meanwhile, GPT-4o demonstrates better performance than LLaVA-34B. 4) By standing upon the shoulders of giants, our proposed \textsc{LoReHM} could achieve superior performance than the `LMM backbones (\ie, LLaVA-34B and GPT-4o) with few-shot ICL prompts', which notably improves over GPT-4o by 2.75\%, 4.40\%, and 2.46\% in terms of macro-averaged F1 score on HarM, FHM, and MAMI. Overall, our \textsc{LoReHM} based on both representative open-source and closed-source LMMs showcases consistent and adaptable performance across all benchmark datasets for harmful meme detection, thanks to its astute discernment of harmful memes in the low-resource few-shot setting.

\subsection{Ablation Study}
We perform ablative studies by adding the paradigms on LMM agents to draw more insightful comparisons among variants of LMMs, as shown in Table~\ref{tab:ablation}. LLaVA-34B and GPT-4o are selected as the representative LMMs from the open-source and closed-source perspectives. We devise five variants of paradigms based on LMM agents for low-resource harmful meme detection: 1) \textit{w/ 0-shot Prompt}: Directly prompt a representative LMM, to infer harmfulness for harmful meme detection; 2) \textit{w/ 50-shot ICL}: Prompt the LMM with 50-shot ICL demonstrations, the similar setting to the third-group baselines in Table~\ref{tab:main_results}; 3) \textit{w/ Relative Sample Augmentation (RSA)}: Augment the LMM agent by looking outward with the label information of the retrieved memes as prior; 4) \textit{w/ Meme Insight Augmentation (MIA)}: Augment the LMM agent by looking inward through mimicking the knowledge-revising behavior of human problem-solving skills to get general insights into the meme harmfulness; 5) \textit{w/ \textsc{LoReHM}}: Our proposed \textsc{LoReHM} based on the full integration of the RSA\&MIA strategies.

We have the following observations: 1) The direct deployment `\textit{w/ 0-shot Prompt}' on LLaVA-34B and GPT-4o struggles since the models are not specifically designed for this task. 2) The `\textit{50-shot ICL}' prompting strategy could effectively enhance the detection performance of LMMs, though the repeated combination of 50-shot examples with test instances incurs significant computational overhead during each inference step. 3) The overall performances of the `\textit{w/ RSA}' and `\textit{w/ MIA}' settings are equally matched even slightly improved when compared with the `\textit{w/ 50-shot ICL}' setting. However, `LLaVA-34B \textit{w/ RSA}' on FHM and MAMI datasets does not perform well, for which we speculate the reason is that LLaVA-34B tends to believe the prior judgment given by the RSA strategy instead of more insight reflection as GPT-4o. 4) Moreover, the full paradigm `\textit{w/ \textsc{LoReHM}}' could yield an absolute performance advantage rather than using either `\textit{w/ RSA}' or `\textit{w/ MIA}' alone, which implies the proposed two strategies are not incremental but complementary each other. On the other hand, the `\textit{w/ \textsc{LoReHM}}' setting achieves excellent and robust performance based on both LLaVA-34B and GPT-4o, which demonstrates that the choice of LMMs is orthogonal to our proposed agentic paradigm that can be easily augmented with newly released stronger LMMs without any other change. 

\begin{figure}[t]
    \centering
    \includegraphics[width=1\linewidth]{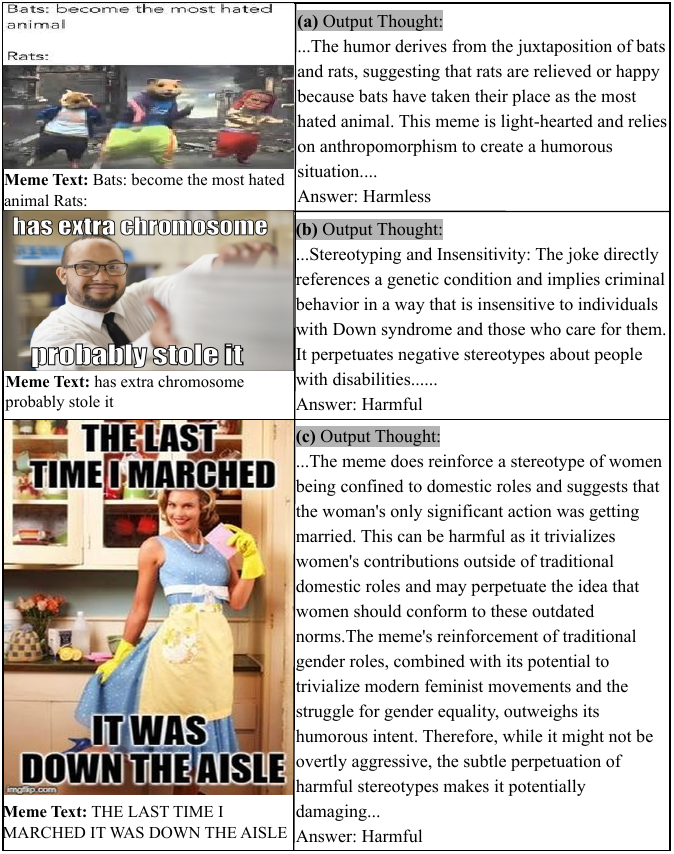}
    \vspace{-0.7cm}
    \caption{Examples of correctly predicted memes in (a) HarM, (b) FHM,  and (c) MAMI datasets. }
    \label{fig:case_study}
    \vspace{-0.4cm}
\end{figure}

\subsection{Case Study}
To better understand how the LMM agent deciphers the test meme samples, we conduct a case study on the output thought of the LMM agent for the correctly predicted samples, where we show important content in the thought and truncate others. as exemplified in Figure~\ref{fig:case_study}.

From the LMM agent's output thought in the natural text, we can observe that:
1) The agent effectively links multimodal information related to the meme text and image using commonsense knowledge. For example, in Figure~\ref{fig:case_study}(a),  ``rats are relieved or happy'' in the thought could be linked to the dancing rats in the image, and ``the juxtaposition of bats and rats'' in the thought could be linked to ``Bats'' and ``Rats'' in the text; In Figure~\ref{fig:case_study}(b), ``Down syndrome'' in the thought could be linked to ``extra chromosome'' in the text; and in terms of Figure~\ref{fig:case_study}(c), ``women being confined to domestic roles'' in the thought could be linked to a woman doing housework in the image.
2) Furthermore, the agent demonstrates advanced reasoning by considering the interplay of multimodal information. In Figure~\ref{fig:case_study}(a), the thought takes into account the cultural context and the potential emotional response of the viewer, ultimately appreciating the light-hearted intent behind the meme's creation; The thought in Figure~\ref{fig:case_study}(b) explores the implications of making light of a serious issue, recognizing the potential harm in perpetuating stereotypes and insensitivity towards individuals with Down syndrome; For Figure~\ref{fig:case_study}(c), the thought scrutinizes the reinforcement of domestic role stereotypes, evaluating the potential for such stereotypes to desensitize viewers to the importance of gender equality. In this way, the rich but implicit correlations between the meme text and image are explained in readable snippets, which can be potentially valuable for aiding human checkers in verifying model predictions in the low-resource setting. We also provide more case studies and error analysis in the Appendix \S\ref{sec:more_cases} and \S\ref{sec:error_analysis}.

\subsection{Effect of Labeled Data Size}

{\setlength{\abovecaptionskip}{-0.1cm}
\setlength{\belowcaptionskip}{-0.1cm}
\begin{figure}[t]
\centering
\subfigure{
\begin{minipage}[t]{0.5\linewidth}
\centering
\scalebox{0.75}{\includegraphics[width=5cm]{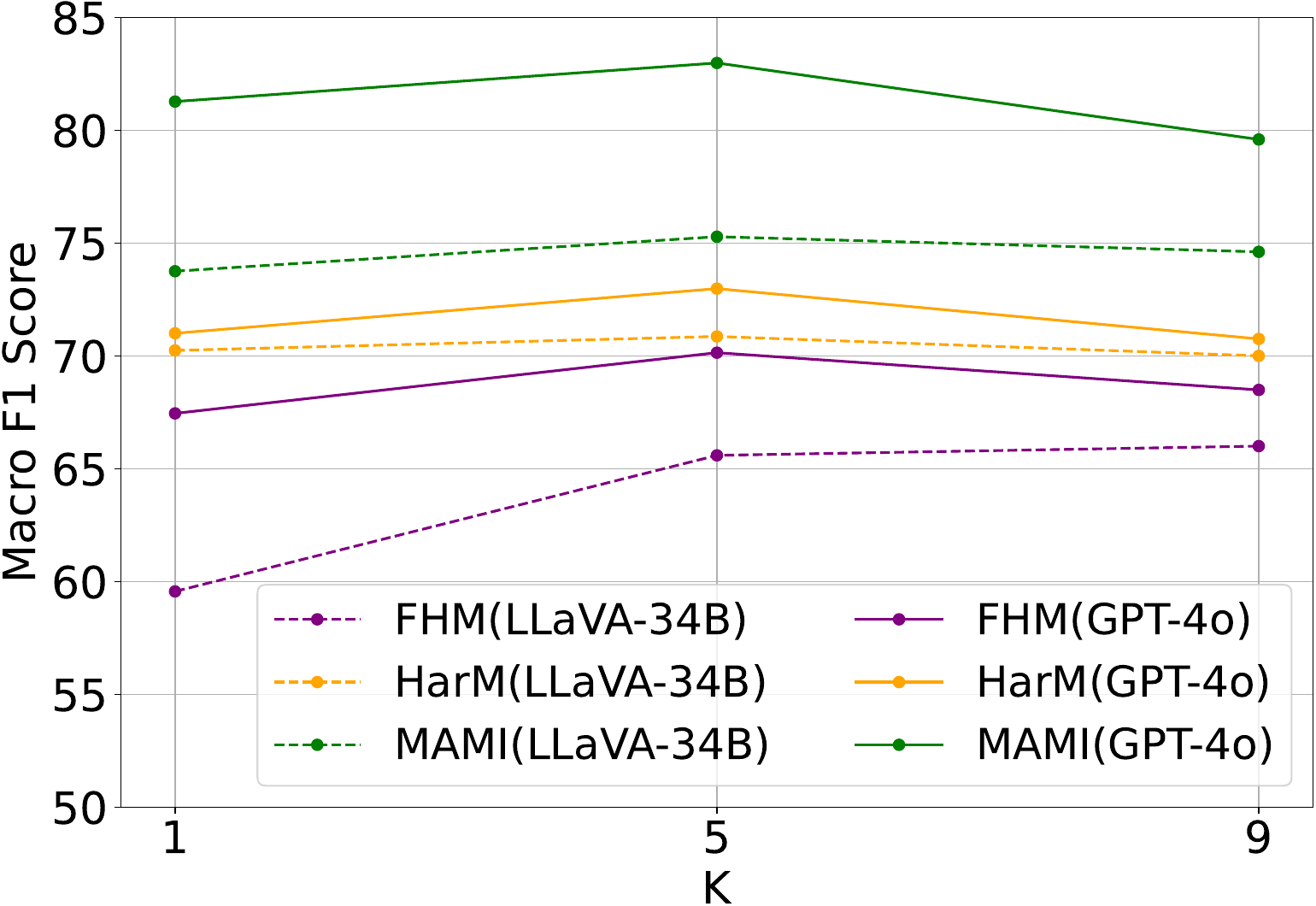}}
\end{minipage}%
}%
\subfigure{
\begin{minipage}[t]{0.5\linewidth}
\centering
\scalebox{0.75}{\includegraphics[width=5cm]{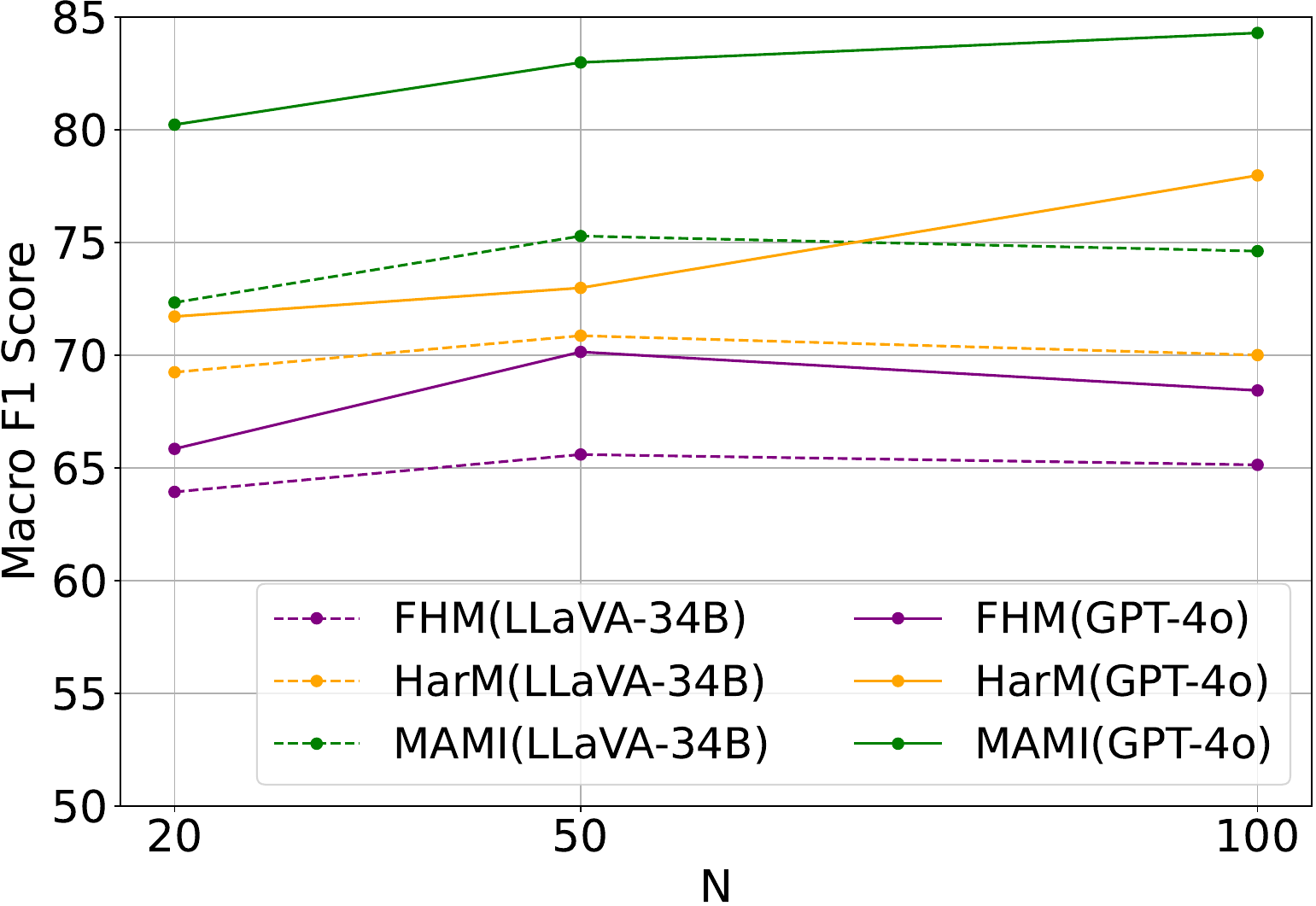}}
\end{minipage}%
}%
\centering
\caption{Effect of $Top_K$ and $N$-shot.}
\label{fig:vis}
\vspace{-0.4cm}
\end{figure}}

To study the effects of the labeled data size in our agency-driven paradigm, we conduct an analysis of performance with $Top_K$ (left) and $N$-shot (right) as shown in Figure~\ref{fig:vis}. We use the control variable method to adjust the values of $K$ and $N$, respectively. It is observed that: 1) As $K$ increases, the gap between different backbone variants decreases. 2) Despite increasing the number $N$ of labeled examples, performance plateaus or declines in some configurations, indicating that harmful meme detection remains challenging in a few-shot framework. This suggests that simply adding more examples is not enough. Innovative approaches and advanced techniques are needed to better leverage few-shot meme examples and capture multimodal subtleties.

\section{Conclusion and Future Work}
In this paper, we delved into the low-resource issue of harmful meme detection with limited few-shot annotations. To this end, we proposed an agency-driven and gradient-free approach, which seamlessly integrated the Relative Sample Augmentation and the Meme Insight Augmentation strategies to strengthen LMMs as a trustworthy agent. Comprehensive experiments and analyses confirm the advantages of our agentic framework. Future efforts aim to enhance our research by further relieving the inherent bias and variation in LMMs.

\section*{Limitations}
There are multiple ways to further improve this work:
\begin{itemize}
    \item Firstly, although harmfulness is defined much broader than hatefulness or misogyny in previous literature~\cite{pramanick2021detecting}, in the future, we would try to incorporate more of the related meme datasets beyond our task to further broaden the boundaries of this low-resource research, such as disinformation~\cite{wang2024mfc}, offensiveness~\cite{suryawanshi2020multimodal}, sarcasm~\cite{lin2024cofipara}, and even code-mixed data~\cite{maity2022multitask}, etc.
    \item Secondly, in this work, we primarily focus on the few-shot setting to address the low-resource issue. We plan to further explore the zero-shot setting, which presents an even more challenging aspect of the low-resource problem. Additionally, it is also crucial for our future research to understand meme data in the context of low-resource domains and languages on social media~\cite{lin2022detect}.
    \item Thirdly, while this work focuses on improving few-shot performance in low-resource harmful meme detection, assessing the quality of the extracted insights remains challenging and inherently qualitative. Given that our agentic framework can generate readable snippets for cognitive-view reasoning, we plan to conduct a systematic study to evaluate and claim explainability. This would constitute another targeted area of research.
    \item Lastly, since this work focuses on the investigation of the agentic memory algorithm for the low-resource harmful meme detection task, we would further explore the work from the perspective of agentic planning algorithms, update our framework by incorporating more emerging LMMs if accessible in the future, and continue to avoid several common deficiencies of existing LMMs, including hallucination, inherent bias, and limited generalization as much as possible.
\end{itemize}

\section*{Ethics Statement}
The purpose of this work is to prevent the spread of harmful meme information and to ensure that people are not subjected to prejudice or racial and gender discrimination. Nevertheless, we are aware of the potential for malicious users to reverse-engineer and create memes that go undetected or misunderstood by AI systems based on \textsc{LoReHM}. This is strongly discouraged and condemned. Intervention with human moderation would be required in order to ensure that this does not occur. Research indicates that evaluating harmful or hateful content can have negative effects. Our proposed agency-driven paradigm with LMMs could generate insightful thought, which can provide human users or checkers with dialectical thinking that allows them to better decode the underlying meaning of memes. Another consideration is the usage of Facebook’s meme dataset; users will have to agree with Facebook’s usage agreement to gain access to the memes. The usage of Facebook’s memes in this study is in accordance with its usage agreement. All the datasets only include memes and do not contain any user information.

\section*{Acknowledgements}
This work is partially supported by National Natural Science Foundation of China Young Scientists Fund (No. 62206233) and 
Hong Kong RGC ECS (No. 22200722).

\bibliography{custom}

\begin{thebibliography}{81}
\providecommand{\natexlab}[1]{#1}

\bibitem[{Alayrac et~al.(2022)Alayrac, Donahue, Luc, Miech, Barr, Hasson, Lenc, Mensch, Millican, Reynolds et~al.}]{alayrac2022flamingo}
Jean-Baptiste Alayrac, Jeff Donahue, Pauline Luc, Antoine Miech, Iain Barr, Yana Hasson, Karel Lenc, Arthur Mensch, Katherine Millican, Malcolm Reynolds, et~al. 2022.
\newblock Flamingo: a visual language model for few-shot learning.
\newblock In \emph{Advances in Neural Information Processing Systems}.

\bibitem[{Asai et~al.(2023)Asai, Wu, Wang, Sil, and Hajishirzi}]{asai2023self}
Akari Asai, Zeqiu Wu, Yizhong Wang, Avirup Sil, and Hannaneh Hajishirzi. 2023.
\newblock Self-rag: Learning to retrieve, generate, and critique through self-reflection.
\newblock In \emph{The Twelfth International Conference on Learning Representations}.

\bibitem[{Awadalla et~al.(2023)Awadalla, Gao, Gardner, Hessel, Hanafy, Zhu, Marathe, Bitton, Gadre, Sagawa et~al.}]{awadalla2023openflamingo}
Anas Awadalla, Irena Gao, Josh Gardner, Jack Hessel, Yusuf Hanafy, Wanrong Zhu, Kalyani Marathe, Yonatan Bitton, Samir Gadre, Shiori Sagawa, et~al. 2023.
\newblock Openflamingo: An open-source framework for training large autoregressive vision-language models.
\newblock \emph{arXiv preprint arXiv:2308.01390}.

\bibitem[{Bai et~al.(2023)Bai, Bai, Yang, Wang, Tan, Wang, Lin, Zhou, and Zhou}]{bai2023qwen}
Jinze Bai, Shuai Bai, Shusheng Yang, Shijie Wang, Sinan Tan, Peng Wang, Junyang Lin, Chang Zhou, and Jingren Zhou. 2023.
\newblock Qwen-vl: A frontier large vision-language model with versatile abilities.
\newblock \emph{arXiv preprint arXiv:2308.12966}.

\bibitem[{Baran(2001)}]{baran2001prologue}
Stanley Baran. 2001.
\newblock Prologue—framing public life: A bridging model for media research.
\newblock \emph{Framing Public Life: Perspectives on Media and Our Understanding of the Social World}, page~7.

\bibitem[{Black et~al.(2022)Black, Biderman, Hallahan, Anthony, Gao, Golding, He, Leahy, McDonell, Phang et~al.}]{black2022gpt}
Sidney Black, Stella Biderman, Eric Hallahan, Quentin Anthony, Leo Gao, Laurence Golding, Horace He, Connor Leahy, Kyle McDonell, Jason Phang, et~al. 2022.
\newblock Gpt-neox-20b: An open-source autoregressive language model.
\newblock In \emph{Proceedings of BigScience Episode\# 5--Workshop on Challenges \& Perspectives in Creating Large Language Models}, pages 95--136.

\bibitem[{Brown et~al.(2020)Brown, Mann, Ryder, Subbiah, Kaplan, Dhariwal, Neelakantan, Shyam, Sastry, Askell et~al.}]{brown2020language}
Tom~B Brown, Benjamin Mann, Nick Ryder, Melanie Subbiah, Jared Kaplan, Prafulla Dhariwal, Arvind Neelakantan, Pranav Shyam, Girish Sastry, Amanda Askell, et~al. 2020.
\newblock Language models are few-shot learners.
\newblock In \emph{Proceedings of the 34th International Conference on Neural Information Processing Systems}, pages 1877--1901.

\bibitem[{Cao et~al.(2023)Cao, Hee, Kuek, Chong, Lee, and Jiang}]{cao2023pro}
Rui Cao, Ming~Shan Hee, Adriel Kuek, Wen-Haw Chong, Roy Ka-Wei Lee, and Jing Jiang. 2023.
\newblock Pro-cap: Leveraging a frozen vision-language model for hateful meme detection.
\newblock In \emph{Proceedings of the 31th ACM international conference on multimedia}.

\bibitem[{Cao et~al.(2022)Cao, Lee, Chong, and Jiang}]{cao2023prompting}
Rui Cao, Roy Ka-Wei Lee, Wen-Haw Chong, and Jing Jiang. 2022.
\newblock Prompting for multimodal hateful meme classification.
\newblock In \emph{Proceedings of the 2022 Conference on Empirical Methods in Natural Language Processing}, pages 321--332.

\bibitem[{Cao et~al.(2024)Cao, Lee, and Jiang}]{cao2024modularized}
Rui Cao, Roy Ka-Wei Lee, and Jing Jiang. 2024.
\newblock Modularized networks for few-shot hateful meme detection.
\newblock \emph{arXiv preprint arXiv:2402.11845}.

\bibitem[{Chen et~al.(2022)Chen, Zhang, Nguyen, Zan, Lin, Lou, and Chen}]{chen2022codet}
Bei Chen, Fengji Zhang, Anh Nguyen, Daoguang Zan, Zeqi Lin, Jian-Guang Lou, and Weizhu Chen. 2022.
\newblock Codet: Code generation with generated tests.
\newblock In \emph{The Eleventh International Conference on Learning Representations}.

\bibitem[{Chen et~al.(2023)Chen, Lin, Schaerli, and Zhou}]{chen2023teaching}
Xinyun Chen, Maxwell Lin, Nathanael Schaerli, and Denny Zhou. 2023.
\newblock Teaching large language models to self-debug.
\newblock In \emph{The 61st Annual Meeting Of The Association For Computational Linguistics}.

\bibitem[{Chowdhery et~al.(2022)Chowdhery, Narang, Devlin, Bosma, Mishra, Roberts, Barham, Chung, Sutton, Gehrmann et~al.}]{chowdhery2022palm}
Aakanksha Chowdhery, Sharan Narang, Jacob Devlin, Maarten Bosma, Gaurav Mishra, Adam Roberts, Paul Barham, Hyung~Won Chung, Charles Sutton, Sebastian Gehrmann, et~al. 2022.
\newblock Palm: Scaling language modeling with pathways.
\newblock \emph{arXiv preprint arXiv:2204.02311}.

\bibitem[{Dai et~al.(2023)Dai, Li, Li, Tiong, Zhao, Wang, Li, Fung, and Hoi}]{Dai2023InstructBLIPTG}
Wenliang Dai, Junnan Li, Dongxu Li, Anthony Meng~Huat Tiong, Junqi Zhao, Weisheng Wang, Boyang~Albert Li, Pascale Fung, and Steven C.~H. Hoi. 2023.
\newblock Instructblip: Towards general-purpose vision-language models with instruction tuning.
\newblock \emph{ArXiv}.

\bibitem[{Das et~al.(2020)Das, Wahi, and Li}]{das2020detecting}
Abhishek Das, Japsimar~Singh Wahi, and Siyao Li. 2020.
\newblock Detecting hate speech in multi-modal memes.
\newblock \emph{arXiv preprint arXiv:2012.14891}.

\bibitem[{Devlin et~al.(2019)Devlin, Chang, Lee, and Toutanova}]{devlin2018bert}
Jacob Devlin, Ming-Wei Chang, Kenton Lee, and Kristina Toutanova. 2019.
\newblock Bert: Pre-training of deep bidirectional transformers for language understanding.
\newblock In \emph{Proceedings of NAACL-HLT}, pages 4171--4186.

\bibitem[{Fersini et~al.(2022)Fersini, Gasparini, Rizzi, Saibene, Chulvi, Rosso, Lees, and Sorensen}]{fersini2022semeval}
Elisabetta Fersini, Francesca Gasparini, Giulia Rizzi, Aurora Saibene, Berta Chulvi, Paolo Rosso, Alyssa Lees, and Jeffrey Sorensen. 2022.
\newblock Semeval-2022 task 5: Multimedia automatic misogyny identification.
\newblock In \emph{Proceedings of the 16th International Workshop on Semantic Evaluation (SemEval-2022)}, pages 533--549.

\bibitem[{Gao et~al.(2023)Gao, Yen, Yu, and Chen}]{gao2023enabling}
Tianyu Gao, Howard Yen, Jiatong Yu, and Danqi Chen. 2023.
\newblock Enabling large language models to generate text with citations.
\newblock In \emph{Proceedings of the 2023 Conference on Empirical Methods in Natural Language Processing}, pages 6465--6488.

\bibitem[{Guu et~al.(2020)Guu, Lee, Tung, Pasupat, and Chang}]{guu2020retrieval}
Kelvin Guu, Kenton Lee, Zora Tung, Panupong Pasupat, and Mingwei Chang. 2020.
\newblock Retrieval augmented language model pre-training.
\newblock In \emph{International conference on machine learning}, pages 3929--3938. PMLR.

\bibitem[{Hee et~al.(2022)Hee, Lee, and Chong}]{hee2022explaining}
Ming~Shan Hee, Roy Ka-Wei Lee, and Wen-Haw Chong. 2022.
\newblock On explaining multimodal hateful meme detection models.
\newblock In \emph{Proceedings of the ACM Web Conference 2022}, pages 3651--3655.

\bibitem[{Hong et~al.(2023)Hong, Zhuge, Chen, Zheng, Cheng, Wang, Zhang, Wang, Yau, Lin et~al.}]{hong2023metagpt}
Sirui Hong, Mingchen Zhuge, Jonathan Chen, Xiawu Zheng, Yuheng Cheng, Jinlin Wang, Ceyao Zhang, Zili Wang, Steven Ka~Shing Yau, Zijuan Lin, et~al. 2023.
\newblock Metagpt: Meta programming for multi-agent collaborative framework.
\newblock In \emph{The Twelfth International Conference on Learning Representations}.

\bibitem[{Hu et~al.(2021)Hu, Wallis, Allen-Zhu, Li, Wang, Wang, Chen et~al.}]{hu2021lora}
Edward~J Hu, Phillip Wallis, Zeyuan Allen-Zhu, Yuanzhi Li, Shean Wang, Lu~Wang, Weizhu Chen, et~al. 2021.
\newblock Lora: Low-rank adaptation of large language models.
\newblock In \emph{International Conference on Learning Representations}.

\bibitem[{Izacard et~al.(2023)Izacard, Lewis, Lomeli, Hosseini, Petroni, Schick, Dwivedi-Yu, Joulin, Riedel, and Grave}]{izacard2023atlas}
Gautier Izacard, Patrick Lewis, Maria Lomeli, Lucas Hosseini, Fabio Petroni, Timo Schick, Jane Dwivedi-Yu, Armand Joulin, Sebastian Riedel, and Edouard Grave. 2023.
\newblock Atlas: Few-shot learning with retrieval augmented language models.
\newblock \emph{Journal of Machine Learning Research}, 24(251):1--43.

\bibitem[{Ji et~al.(2023)Ji, Ren, and Naseem}]{ji2023identifying}
Junhui Ji, Wei Ren, and Usman Naseem. 2023.
\newblock Identifying creative harmful memes via prompt based approach.
\newblock In \emph{Proceedings of the ACM Web Conference 2023}, pages 3868--3872.

\bibitem[{Jiang et~al.(2023)Jiang, Xu, Gao, Sun, Liu, Dwivedi-Yu, Yang, Callan, and Neubig}]{jiang2023active}
Zhengbao Jiang, Frank~F Xu, Luyu Gao, Zhiqing Sun, Qian Liu, Jane Dwivedi-Yu, Yiming Yang, Jamie Callan, and Graham Neubig. 2023.
\newblock Active retrieval augmented generation.
\newblock In \emph{Proceedings of the 2023 Conference on Empirical Methods in Natural Language Processing}, pages 7969--7992.

\bibitem[{Kiela et~al.(2019)Kiela, Bhooshan, Firooz, Perez, and Testuggine}]{kiela2019supervised}
Douwe Kiela, Suvrat Bhooshan, Hamed Firooz, Ethan Perez, and Davide Testuggine. 2019.
\newblock Supervised multimodal bitransformers for classifying images and text.
\newblock \emph{arXiv preprint arXiv:1909.02950}.

\bibitem[{Kiela et~al.(2020)Kiela, Firooz, Mohan, Goswami, Singh, Ringshia, and Testuggine}]{kiela2020hateful}
Douwe Kiela, Hamed Firooz, Aravind Mohan, Vedanuj Goswami, Amanpreet Singh, Pratik Ringshia, and Davide Testuggine. 2020.
\newblock The hateful memes challenge: detecting hate speech in multimodal memes.
\newblock In \emph{Proceedings of the 34th International Conference on Neural Information Processing Systems}, pages 2611--2624.

\bibitem[{Kirillov et~al.(2023)Kirillov, Mintun, Ravi, Mao, Rolland, Gustafson, Xiao, Whitehead, Berg, Lo et~al.}]{kirillov2023segment}
Alexander Kirillov, Eric Mintun, Nikhila Ravi, Hanzi Mao, Chloe Rolland, Laura Gustafson, Tete Xiao, Spencer Whitehead, Alexander~C Berg, Wan-Yen Lo, et~al. 2023.
\newblock Segment anything.
\newblock \emph{arXiv preprint arXiv:2304.02643}.

\bibitem[{Kojima et~al.(2022)Kojima, Gu, Reid, Matsuo, and Iwasawa}]{kojima2022large}
Takeshi Kojima, Shixiang~Shane Gu, Machel Reid, Yutaka Matsuo, and Yusuke Iwasawa. 2022.
\newblock Large language models are zero-shot reasoners.
\newblock In \emph{ICML 2022 Workshop on Knowledge Retrieval and Language Models}.

\bibitem[{Kuang et~al.(2021)Kuang, Sun, Li, Yue, Lin, Chen, Wei, Zhu, Gao, Zhang et~al.}]{kuang2021mmocr}
Zhanghui Kuang, Hongbin Sun, Zhizhong Li, Xiaoyu Yue, Tsui~Hin Lin, Jianyong Chen, Huaqiang Wei, Yiqin Zhu, Tong Gao, Wenwei Zhang, et~al. 2021.
\newblock Mmocr: a comprehensive toolbox for text detection, recognition and understanding.
\newblock In \emph{Proceedings of the 29th ACM International Conference on Multimedia}, pages 3791--3794.

\bibitem[{Lee et~al.(2021)Lee, Cao, Fan, Jiang, and Chong}]{lee2021disentangling}
Roy Ka-Wei Lee, Rui Cao, Ziqing Fan, Jing Jiang, and Wen-Haw Chong. 2021.
\newblock Disentangling hate in online memes.
\newblock In \emph{Proceedings of the 29th ACM International Conference on Multimedia}, pages 5138--5147.

\bibitem[{Lewis et~al.(2020)Lewis, Perez, Piktus, Petroni, Karpukhin, Goyal, K{\"u}ttler, Lewis, Yih, Rockt{\"a}schel et~al.}]{lewis2020retrieval}
Patrick Lewis, Ethan Perez, Aleksandra Piktus, Fabio Petroni, Vladimir Karpukhin, Naman Goyal, Heinrich K{\"u}ttler, Mike Lewis, Wen-tau Yih, Tim Rockt{\"a}schel, et~al. 2020.
\newblock Retrieval-augmented generation for knowledge-intensive nlp tasks.
\newblock In \emph{Proceedings of the 34th International Conference on Neural Information Processing Systems}, pages 9459--9474.

\bibitem[{Lin et~al.(2022{\natexlab{a}})Lin, Chen, Ma, Yang, and Chen}]{lin2022amif}
Hongzhan Lin, Liangliang Chen, Jing Ma, Zhiwei Yang, and Guang Chen. 2022{\natexlab{a}}.
\newblock Amif: A hybrid model for improving fact checking in product question answering.
\newblock In \emph{2022 International Joint Conference on Neural Networks (IJCNN)}, pages 1--8. IEEE.

\bibitem[{Lin et~al.(2024{\natexlab{a}})Lin, Chen, Luo, Cheng, Ma, and Chen}]{lin2024cofipara}
Hongzhan Lin, Zixin Chen, Ziyang Luo, Mingfei Cheng, Jing Ma, and Guang Chen. 2024{\natexlab{a}}.
\newblock Cofipara: A coarse-to-fine paradigm for multimodal sarcasm target identification with large multimodal models.
\newblock \emph{arXiv preprint arXiv:2405.00390}.

\bibitem[{Lin et~al.(2024{\natexlab{b}})Lin, Luo, Gao, Ma, Wang, and Yang}]{lin2024explainable}
Hongzhan Lin, Ziyang Luo, Wei Gao, Jing Ma, Bo~Wang, and Ruichao Yang. 2024{\natexlab{b}}.
\newblock Towards explainable harmful meme detection through multimodal debate between large language models.
\newblock In \emph{The ACM Web Conference 2024}, Singapore.

\bibitem[{Lin et~al.(2023{\natexlab{a}})Lin, Luo, Ma, and Chen}]{lin2023beneath}
Hongzhan Lin, Ziyang Luo, Jing Ma, and Long Chen. 2023{\natexlab{a}}.
\newblock Beneath the surface: Unveiling harmful memes with multimodal reasoning distilled from large language models.
\newblock In \emph{The 2023 Conference on Empirical Methods in Natural Language Processing}.

\bibitem[{Lin et~al.(2024{\natexlab{c}})Lin, Luo, Wang, Yang, and Ma}]{lin2024goat}
Hongzhan Lin, Ziyang Luo, Bo~Wang, Ruichao Yang, and Jing Ma. 2024{\natexlab{c}}.
\newblock Goat-bench: Safety insights to large multimodal models through meme-based social abuse.
\newblock \emph{arXiv preprint arXiv:2401.01523}.

\bibitem[{Lin et~al.(2022{\natexlab{b}})Lin, Ma, Chen, Yang, Cheng, and Guang}]{lin2022detect}
Hongzhan Lin, Jing Ma, Liangliang Chen, Zhiwei Yang, Mingfei Cheng, and Chen Guang. 2022{\natexlab{b}}.
\newblock Detect rumors in microblog posts for low-resource domains via adversarial contrastive learning.
\newblock In \emph{Findings of the Association for Computational Linguistics: NAACL 2022}, pages 2543--2556.

\bibitem[{Lin et~al.(2021)Lin, Yan, and Chen}]{lin2021boosting}
Hongzhan Lin, Yuanmeng Yan, and Guang Chen. 2021.
\newblock Boosting low-resource intent detection with in-scope prototypical networks.
\newblock In \emph{ICASSP 2021-2021 IEEE International Conference on Acoustics, Speech and Signal Processing (ICASSP)}, pages 7623--7627. IEEE.

\bibitem[{Lin et~al.(2023{\natexlab{b}})Lin, Yi, Ma, Jiang, Luo, Shi, and Liu}]{lin2023zero}
Hongzhan Lin, Pengyao Yi, Jing Ma, Haiyun Jiang, Ziyang Luo, Shuming Shi, and Ruifang Liu. 2023{\natexlab{b}}.
\newblock Zero-shot rumor detection with propagation structure via prompt learning.
\newblock In \emph{Proceedings of the AAAI Conference on Artificial Intelligence}, volume~37, pages 5213--5221.

\bibitem[{Lippe et~al.(2020)Lippe, Holla, Chandra, Rajamanickam, Antoniou, Shutova, and Yannakoudakis}]{lippe2020multimodal}
Phillip Lippe, Nithin Holla, Shantanu Chandra, Santhosh Rajamanickam, Georgios Antoniou, Ekaterina Shutova, and Helen Yannakoudakis. 2020.
\newblock A multimodal framework for the detection of hateful memes.
\newblock \emph{arXiv preprint arXiv:2012.12871}.

\bibitem[{Liu et~al.(2024)Liu, Li, Li, Li, Zhang, Shen, and Lee}]{liu2024llava}
Haotian Liu, Chunyuan Li, Yuheng Li, Bo~Li, Yuanhan Zhang, Sheng Shen, and Yong~Jae Lee. 2024.
\newblock Llava-next: Improved reasoning, ocr, and world knowledge (january 2024).
\newblock \emph{URL https://llava-vl. github. io/blog/2024-01-30-llava-next}, 1(8).

\bibitem[{Liu et~al.(2023{\natexlab{a}})Liu, Li, Wu, and Lee}]{liu2023visual}
Haotian Liu, Chunyuan Li, Qingyang Wu, and Yong~Jae Lee. 2023{\natexlab{a}}.
\newblock Visual instruction tuning.
\newblock \emph{arXiv preprint arXiv:2304.08485}.

\bibitem[{Liu et~al.(2022)Liu, Shen, Zhang, Dolan, Carin, and Chen}]{liu2022makes}
Jiachang Liu, Dinghan Shen, Yizhe Zhang, William~B Dolan, Lawrence Carin, and Weizhu Chen. 2022.
\newblock What makes good in-context examples for gpt-3?
\newblock In \emph{Proceedings of Deep Learning Inside Out (DeeLIO 2022): The 3rd Workshop on Knowledge Extraction and Integration for Deep Learning Architectures}, pages 100--114.

\bibitem[{Liu et~al.(2023{\natexlab{b}})Liu, Yu, Zhang, Xu, Lei, Lai, Gu, Ding, Men, Yang et~al.}]{liu2023agentbench}
Xiao Liu, Hao Yu, Hanchen Zhang, Yifan Xu, Xuanyu Lei, Hanyu Lai, Yu~Gu, Hangliang Ding, Kaiwen Men, Kejuan Yang, et~al. 2023{\natexlab{b}}.
\newblock Agentbench: Evaluating llms as agents.
\newblock In \emph{The Twelfth International Conference on Learning Representations}.

\bibitem[{Luo et~al.(2023)Luo, Xu, Zhao, Sun, Geng, Hu, Tao, Ma, Lin, and Jiang}]{luo2023wizardcoder}
Ziyang Luo, Can Xu, Pu~Zhao, Qingfeng Sun, Xiubo Geng, Wenxiang Hu, Chongyang Tao, Jing Ma, Qingwei Lin, and Daxin Jiang. 2023.
\newblock Wizardcoder: Empowering code large language models with evol-instruct.
\newblock In \emph{The Twelfth International Conference on Learning Representations}.

\bibitem[{Madaan et~al.(2023)Madaan, Tandon, Gupta, Hallinan, Gao, Wiegreffe, Alon, Dziri, Prabhumoye, Yang et~al.}]{madaan2023self}
Aman Madaan, Niket Tandon, Prakhar Gupta, Skyler Hallinan, Luyu Gao, Sarah Wiegreffe, Uri Alon, Nouha Dziri, Shrimai Prabhumoye, Yiming Yang, et~al. 2023.
\newblock Self-refine: Iterative refinement with self-feedback.
\newblock In \emph{Thirty-seventh Conference on Neural Information Processing Systems}.

\bibitem[{Maity et~al.(2022)Maity, Jha, Saha, and Bhattacharyya}]{maity2022multitask}
Krishanu Maity, Prince Jha, Sriparna Saha, and Pushpak Bhattacharyya. 2022.
\newblock A multitask framework for sentiment, emotion and sarcasm aware cyberbullying detection from multi-modal code-mixed memes.
\newblock In \emph{Proceedings of the 45th International ACM SIGIR Conference on Research and Development in Information Retrieval}, pages 1739--1749.

\bibitem[{Mu et~al.(2023)Mu, Zhang, Hu, Wang, Ding, Jin, Wang, Dai, Qiao, and Luo}]{mu2023embodiedgpt}
Yao Mu, Qinglong Zhang, Mengkang Hu, Wenhai Wang, Mingyu Ding, Jun Jin, Bin Wang, Jifeng Dai, Yu~Qiao, and Ping Luo. 2023.
\newblock Embodiedgpt: Vision-language pre-training via embodied chain of thought.
\newblock In \emph{Thirty-seventh Conference on Neural Information Processing Systems}.

\bibitem[{Muennighoff(2020)}]{muennighoff2020vilio}
Niklas Muennighoff. 2020.
\newblock Vilio: State-of-the-art visio-linguistic models applied to hateful memes.
\newblock \emph{arXiv preprint arXiv:2012.07788}.

\bibitem[{OpenAI(2023)}]{OpenAI2023GPT4TR}
OpenAI. 2023.
\newblock \href {https://api.semanticscholar.org/CorpusID:257532815} {Gpt-4 technical report}.
\newblock \emph{ArXiv}, abs/2303.08774.

\bibitem[{Ouyang et~al.(2022)Ouyang, Wu, Jiang, Almeida, Wainwright, Mishkin, Zhang, Agarwal, Slama, Ray et~al.}]{ouyang2022training}
Long Ouyang, Jeffrey Wu, Xu~Jiang, Diogo Almeida, Carroll Wainwright, Pamela Mishkin, Chong Zhang, Sandhini Agarwal, Katarina Slama, Alex Ray, et~al. 2022.
\newblock Training language models to follow instructions with human feedback.
\newblock \emph{Advances in Neural Information Processing Systems}, 35:27730--27744.

\bibitem[{Pramanick et~al.(2021{\natexlab{a}})Pramanick, Dimitrov, Mukherjee, Sharma, Akhtar, Nakov, and Chakraborty}]{pramanick2021detecting}
Shraman Pramanick, Dimitar Dimitrov, Rituparna Mukherjee, Shivam Sharma, Md~Shad Akhtar, Preslav Nakov, and Tanmoy Chakraborty. 2021{\natexlab{a}}.
\newblock Detecting harmful memes and their targets.
\newblock In \emph{Findings of the Association for Computational Linguistics: ACL-IJCNLP 2021}, pages 2783--2796.

\bibitem[{Pramanick et~al.(2021{\natexlab{b}})Pramanick, Sharma, Dimitrov, Akhtar, Nakov, and Chakraborty}]{pramanick2021momenta}
Shraman Pramanick, Shivam Sharma, Dimitar Dimitrov, Md~Shad Akhtar, Preslav Nakov, and Tanmoy Chakraborty. 2021{\natexlab{b}}.
\newblock Momenta: A multimodal framework for detecting harmful memes and their targets.
\newblock In \emph{Findings of the Association for Computational Linguistics: EMNLP 2021}, pages 4439--4455.

\bibitem[{Qian et~al.(2023)Qian, Cong, Yang, Chen, Su, Xu, Liu, and Sun}]{qian2023communicative}
Chen Qian, Xin Cong, Cheng Yang, Weize Chen, Yusheng Su, Juyuan Xu, Zhiyuan Liu, and Maosong Sun. 2023.
\newblock Communicative agents for software development.
\newblock \emph{arXiv preprint arXiv:2307.07924}.

\bibitem[{Radford et~al.(2021)Radford, Kim, Hallacy, Ramesh, Goh, Agarwal, Sastry, Askell, Mishkin, Clark et~al.}]{radford2021learning}
Alec Radford, Jong~Wook Kim, Chris Hallacy, Aditya Ramesh, Gabriel Goh, Sandhini Agarwal, Girish Sastry, Amanda Askell, Pamela Mishkin, Jack Clark, et~al. 2021.
\newblock Learning transferable visual models from natural language supervision.
\newblock In \emph{International conference on machine learning}, pages 8748--8763.

\bibitem[{Ram et~al.(2023)Ram, Levine, Dalmedigos, Muhlgay, Shashua, Leyton-Brown, and Shoham}]{ram2023context}
Ori Ram, Yoav Levine, Itay Dalmedigos, Dor Muhlgay, Amnon Shashua, Kevin Leyton-Brown, and Yoav Shoham. 2023.
\newblock In-context retrieval-augmented language models.
\newblock \emph{Transactions of the Association for Computational Linguistics}, 11:1316--1331.

\bibitem[{Ren et~al.(2016)Ren, He, Girshick, and Sun}]{ren2016faster}
Shaoqing Ren, Kaiming He, Ross Girshick, and Jian Sun. 2016.
\newblock Faster r-cnn: Towards real-time object detection with region proposal networks.
\newblock \emph{IEEE Transactions on Pattern Analysis and Machine Intelligence}, 39(6):1137--1149.

\bibitem[{Sandulescu(2020)}]{sandulescu2020detecting}
Vlad Sandulescu. 2020.
\newblock Detecting hateful memes using a multimodal deep ensemble.
\newblock \emph{arXiv preprint arXiv:2012.13235}.

\bibitem[{Schick et~al.(2023)Schick, Dwivedi-Yu, Dessi, Raileanu, Lomeli, Hambro, Zettlemoyer, Cancedda, and Scialom}]{schick2023toolformer}
Timo Schick, Jane Dwivedi-Yu, Roberto Dessi, Roberta Raileanu, Maria Lomeli, Eric Hambro, Luke Zettlemoyer, Nicola Cancedda, and Thomas Scialom. 2023.
\newblock Toolformer: Language models can teach themselves to use tools.
\newblock In \emph{Thirty-seventh Conference on Neural Information Processing Systems}.

\bibitem[{Sharma et~al.(2022)Sharma, Alam, Akhtar, Dimitrov, Da~San~Martino, Firooz, Halevy, Silvestri, Nakov, and Chakraborty}]{sharma2022detecting}
Shivam Sharma, Firoj Alam, Md~Shad Akhtar, Dimitar Dimitrov, Giovanni Da~San~Martino, Hamed Firooz, Alon Halevy, Fabrizio Silvestri, Preslav Nakov, and Tanmoy Chakraborty. 2022.
\newblock Detecting and understanding harmful memes: A survey.
\newblock In \emph{Proceedings of the Thirty-First International Joint Conference on Artificial Intelligence}, pages 5597--5606.

\bibitem[{Shen et~al.(2023)Shen, Song, Tan, Li, Lu, and Zhuang}]{shen2023hugginggpt}
Yongliang Shen, Kaitao Song, Xu~Tan, Dongsheng Li, Weiming Lu, and Yueting Zhuang. 2023.
\newblock Hugginggpt: Solving ai tasks with chatgpt and its friends in hugging face.
\newblock In \emph{Thirty-seventh Conference on Neural Information Processing Systems}.

\bibitem[{Shinn et~al.(2024)Shinn, Cassano, Gopinath, Narasimhan, and Yao}]{shinn2024reflexion}
Noah Shinn, Federico Cassano, Ashwin Gopinath, Karthik Narasimhan, and Shunyu Yao. 2024.
\newblock Reflexion: Language agents with verbal reinforcement learning.
\newblock \emph{Advances in Neural Information Processing Systems}, 36.

\bibitem[{Sun et~al.(2023)Sun, Zhuang, Kong, Dai, and Zhang}]{sun2023adaplanner}
Haotian Sun, Yuchen Zhuang, Lingkai Kong, Bo~Dai, and Chao Zhang. 2023.
\newblock Adaplanner: Adaptive planning from feedback with language models.
\newblock In \emph{Thirty-seventh Conference on Neural Information Processing Systems}.

\bibitem[{Suryawanshi et~al.(2020)Suryawanshi, Chakravarthi, Arcan, and Buitelaar}]{suryawanshi2020multimodal}
Shardul Suryawanshi, Bharathi~Raja Chakravarthi, Mihael Arcan, and Paul Buitelaar. 2020.
\newblock Multimodal meme dataset (multioff) for identifying offensive content in image and text.
\newblock In \emph{Proceedings of the second workshop on trolling, aggression and cyberbullying}, pages 32--41.

\bibitem[{Team et~al.(2023)Team, Anil, Borgeaud, Wu, Alayrac, Yu, Soricut, Schalkwyk, Dai, Hauth et~al.}]{team2023gemini}
Gemini Team, Rohan Anil, Sebastian Borgeaud, Yonghui Wu, Jean-Baptiste Alayrac, Jiahui Yu, Radu Soricut, Johan Schalkwyk, Andrew~M Dai, Anja Hauth, et~al. 2023.
\newblock Gemini: A family of highly capable multimodal models.
\newblock \emph{arXiv preprint arXiv:2312.11805}.

\bibitem[{Touvron et~al.(2023{\natexlab{a}})Touvron, Lavril, Izacard, Martinet, Lachaux, Lacroix, Rozi{\`e}re, Goyal, Hambro, Azhar et~al.}]{touvron2023llama}
Hugo Touvron, Thibaut Lavril, Gautier Izacard, Xavier Martinet, Marie-Anne Lachaux, Timoth{\'e}e Lacroix, Baptiste Rozi{\`e}re, Naman Goyal, Eric Hambro, Faisal Azhar, et~al. 2023{\natexlab{a}}.
\newblock Llama: Open and efficient foundation language models.
\newblock \emph{arXiv preprint arXiv:2302.13971}.

\bibitem[{Touvron et~al.(2023{\natexlab{b}})Touvron, Martin, Stone, Albert, Almahairi, Babaei, Bashlykov, Batra, Bhargava, Bhosale et~al.}]{touvron2023llama2}
Hugo Touvron, Louis Martin, Kevin Stone, Peter Albert, Amjad Almahairi, Yasmine Babaei, Nikolay Bashlykov, Soumya Batra, Prajjwal Bhargava, Shruti Bhosale, et~al. 2023{\natexlab{b}}.
\newblock Llama 2: Open foundation and fine-tuned chat models.
\newblock \emph{arXiv preprint arXiv:2307.09288}.

\bibitem[{Velioglu and Rose(2020)}]{velioglu2020detecting}
Riza Velioglu and Jewgeni Rose. 2020.
\newblock Detecting hate speech in memes using multimodal deep learning approaches: Prize-winning solution to hateful memes challenge.
\newblock \emph{arXiv preprint arXiv:2012.12975}.

\bibitem[{Wang et~al.(2023{\natexlab{a}})Wang, Xie, Jiang, Mandlekar, Xiao, Zhu, Fan, and Anandkumar}]{wang2023voyager}
Guanzhi Wang, Yuqi Xie, Yunfan Jiang, Ajay Mandlekar, Chaowei Xiao, Yuke Zhu, Linxi Fan, and Anima Anandkumar. 2023{\natexlab{a}}.
\newblock Voyager: An open-ended embodied agent with large language models.
\newblock In \emph{Intrinsically-Motivated and Open-Ended Learning Workshop@ NeurIPS2023}.

\bibitem[{Wang et~al.(2024)Wang, Lin, Luo, Ye, Chen, and Ma}]{wang2024mfc}
Shengkang Wang, Hongzhan Lin, Ziyang Luo, Zhen Ye, Guang Chen, and Jing Ma. 2024.
\newblock Mfc-bench: Benchmarking multimodal fact-checking with large vision-language models.
\newblock \emph{arXiv preprint arXiv:2406.11288}.

\bibitem[{Wang et~al.(2023{\natexlab{b}})Wang, Lv, Yu, Hong, Qi, Wang, Ji, Yang, Zhao, Song et~al.}]{wang2023cogvlm}
Weihan Wang, Qingsong Lv, Wenmeng Yu, Wenyi Hong, Ji~Qi, Yan Wang, Junhui Ji, Zhuoyi Yang, Lei Zhao, Xixuan Song, et~al. 2023{\natexlab{b}}.
\newblock Cogvlm: Visual expert for pretrained language models.
\newblock \emph{arXiv preprint arXiv:2311.03079}.

\bibitem[{Wang et~al.(2023{\natexlab{c}})Wang, Li, Sun, and Liu}]{wang2023self}
Yile Wang, Peng Li, Maosong Sun, and Yang Liu. 2023{\natexlab{c}}.
\newblock Self-knowledge guided retrieval augmentation for large language models.
\newblock In \emph{Findings of the Association for Computational Linguistics: EMNLP 2023}, pages 10303--10315.

\bibitem[{Yang et~al.(2023)Yang, Li, Lin, Wang, Lin, Liu, and Wang}]{yang2023dawn}
Zhengyuan Yang, Linjie Li, Kevin Lin, Jianfeng Wang, Chung-Ching Lin, Zicheng Liu, and Lijuan Wang. 2023.
\newblock The dawn of lmms: Preliminary explorations with gpt-4v (ision).
\newblock \emph{arXiv preprint arXiv:2309.17421}, 9(1).

\bibitem[{Yao et~al.(2022)Yao, Zhao, Yu, Du, Shafran, Narasimhan, and Cao}]{yao2022react}
Shunyu Yao, Jeffrey Zhao, Dian Yu, Nan Du, Izhak Shafran, Karthik~R Narasimhan, and Yuan Cao. 2022.
\newblock React: Synergizing reasoning and acting in language models.
\newblock In \emph{The Eleventh International Conference on Learning Representations}.

\bibitem[{Zeng et~al.(2022)Zeng, Liu, Du, Wang, Lai, Ding, Yang, Xu, Zheng, Xia et~al.}]{zeng2022glm}
Aohan Zeng, Xiao Liu, Zhengxiao Du, Zihan Wang, Hanyu Lai, Ming Ding, Zhuoyi Yang, Yifan Xu, Wendi Zheng, Xiao Xia, et~al. 2022.
\newblock Glm-130b: An open bilingual pre-trained model.
\newblock In \emph{The Eleventh International Conference on Learning Representations}.

\bibitem[{Zhang et~al.(2022)Zhang, Roller, Goyal, Artetxe, Chen, Chen, Dewan, Diab, Li, Lin et~al.}]{zhang2022opt}
Susan Zhang, Stephen Roller, Naman Goyal, Mikel Artetxe, Moya Chen, Shuohui Chen, Christopher Dewan, Mona Diab, Xian Li, Xi~Victoria Lin, et~al. 2022.
\newblock Opt: Open pre-trained transformer language models.
\newblock \emph{arXiv preprint arXiv:2205.01068}.

\bibitem[{Zhao et~al.(2024)Zhao, Huang, Xu, Lin, Liu, and Huang}]{zhao2024expel}
Andrew Zhao, Daniel Huang, Quentin Xu, Matthieu Lin, Yong-Jin Liu, and Gao Huang. 2024.
\newblock Expel: Llm agents are experiential learners.
\newblock In \emph{Proceedings of the AAAI Conference on Artificial Intelligence}, volume~38, pages 19632--19642.

\bibitem[{Zhou et~al.(2021)Zhou, Chen, and Yang}]{zhou2021multimodal}
Yi~Zhou, Zhenhao Chen, and Huiyuan Yang. 2021.
\newblock Multimodal learning for hateful memes detection.
\newblock In \emph{2021 IEEE International Conference on Multimedia \& Expo Workshops (ICMEW)}, pages 1--6. IEEE.

\bibitem[{Zhu et~al.(2022)Zhu, Lee, and Chong}]{zhu2022multimodal}
Jiawen Zhu, Roy Ka-Wei Lee, and Wen~Haw Chong. 2022.
\newblock Multimodal zero-shot hateful meme detection.
\newblock In \emph{14th ACM Web Science Conference 2022}, pages 382--389.

\bibitem[{Zhu(2020)}]{zhu2020enhance}
Ron Zhu. 2020.
\newblock Enhance multimodal transformer with external label and in-domain pretrain: Hateful meme challenge winning solution.
\newblock \emph{arXiv preprint arXiv:2012.08290}.

\end{thebibliography}

\appendix

\section{Datasets}
\label{sec:datasets}
\begin{table}[t] \small
\centering
\resizebox{0.35\textwidth}{!}{\begin{tabular}{@{}cccc@{}}
\toprule
\multirow{2}{*}{Datasets} & \multicolumn{2}{c}{Test} \\
                          & \#harmful   & \#harmless   \\ \midrule
HarM                    & 124        & 230         \\
FHM                       & 250        & 250         \\
MAMI                    & 500        & 500         \\\bottomrule
\end{tabular}}
\caption{Statistics of test sets.}
\label{tab:statistics}
\end{table}

The detailed statistics for the original test splits of the three datasets are shown in Table~\ref{tab:statistics}.

\section{Baselines}
\label{sec:baselines}
We compare \textsc{LoReHM} with several state-of-the-art (SoTA) systems for low-resource harmful meme detection: 1) \textsf{PromptHate}~\cite{cao2023prompting}: a prompt learning approach that concatenates the meme text and the image caption as the prompt for masked language modeling; 2) \textsf{\textsc{Mr.Harm}}~\cite{lin2023beneath}: a two-stage framework that distills multimodal reasoning knowledge from LLMs for harmfulness inference;
3) \textsf{Pro-Cap}~\cite{cao2023pro}: a caption-enhanced version of PromptHate, by leveraging pre-trained vision-language models with probing queries, to improve the image caption in the text prompt; 4) \textsf{OPT-30B}~\cite{zhang2022opt}: an early and representative large language model with the in-context learning ability, widely recognized stand-in for GPT-3~\cite{brown2020language}; 5) \textsf{OpenFlamingo-9B}~\cite{awadalla2023openflamingo}: an open-source replication of Flamingo models~\cite{alayrac2022flamingo} that enhances pre-trained, frozen language models by enabling them to cross-attend to the outputs of a frozen vision encoder during the next token prediction; 6) \textsf{Mod-HATE}~\cite{cao2024modularized}: a modularized networks for low-resource harmful meme detection, which train a set of modules capable of relevant tasks and learn a composition of modules with the few-shot examples; 7) \textsf{LLaVA-34B}~\cite{liu2023visual}: an enhanced version of LLaVA, with improved reasoning, OCR, and world knowledge capabilities; 8) \textsf{GPT-4o}~\cite{OpenAI2023GPT4TR}: an optimized version of the proprietary GPT-4 architecture developed by OpenAI, which includes capabilities for processing multiple modalities; 9) \textsf{\textsc{LoReHM} (*)}: Our proposed agentic approach for low-resource harmful meme detection, based on LLaVA-34B and GPT-4o. We use the accuracy and macro-averaged F1 (dominant) scores as the evaluation metrics, where the macro-averaged F1 score is utilized as the monitor to select the best model since it could capture competitive performance beyond the majority class.

\section{Implementation Details}
\label{sec:ID}
To demonstrate the generalizability of our framework, we utilize the representative LMMs LLaVA-34B and GPT-4o as the LMM agent from both the open-source and closed-source perspectives. Specifically, we implement the ``llava-v1.6-34b'' and ``gpt-4o-2024-05-13'' versions for LLaVA-34B and GPT-4o, respectively. Note that the choice of LMMs is orthogonal to our proposed paradigm, which can be easily replaced by newly stronger LMMs without further modification. To make our results reproducible, we set the temperature as 0 without any sampling mechanism. The frozen pre-trained vision and text Transformer encoders are implemented as CLIP~\cite{radford2021learning} with the specific version ``ViT-L/14@336px''. To choose the trade-off parameters $\alpha$ and $\beta$, we conducted a grid search within the range [0,1] and set $\alpha$ and $\beta$ as 0.2 and 0.8, respectively.
For the experiments, all the baselines are tested in the 50-shot setting with balanced classes. In the context of few-shot learning, evaluations can exhibit high variability due to the selection of meme sample examples. To mitigate this, we generate multiple few-shot reference sets using different random seeds for a more reliable few-shot performance evaluation. We create five sets of few-shot examples, each with a different random seed, for each 50-shot setting. Consequently, we report the average accuracy and macro-averaged F1 scores computed over the test set, following model optimization based on these various few-shot samples. As our approach is gradient-free, there are no training parameters. 

We use the released source codes to reproduce PromptHate, \textsc{Mr.Harm}, Pro-Cap, and Mod-HATE. To support the in-context learning of the LLM and/or LMM baselines, we need to first convert the meme's image into an acceptable textual input, because the current LLMs and/or LMMs do not support 50 images as input to have a good performance. It is also a limitation of existing LMMs. We first in-paint the memes by combining MMOCR~\cite{kuang2021mmocr} with SAM~\cite{kirillov2023segment} to extract the text and pure image in memes. Then we apply LLaVA to generate textual descriptions about the dominant objects or events in the memes' image. To ensure the reproducibility of the LLM and/or LMM baselines, we also set the temperature as 0 without any sampling mechanism.

In our approach, to better utilize the preliminary prediction $\mathcal{P}$ provided by Relative Sample Augmentation, we design the prompt template as:

``\textit{A classifier that can identify common features among multiple memes has labeled this meme as \{$\mathcal{P}$\}, Please review the classifier's judgment carefully and use your extensive knowledge to analyze and understand this meme before providing your final verdict. If you disagree with the classifier's judgment, you must provide exceptionally thorough and persuasive reasons.}''

For the insight set $\mathcal{E}_n$ provided by Meme Insight Augmentation, we sequentially concatenate each insight. Subsequently, we concatenate these two parts with the $X_\text{COT}$, resulting in the final prompt input for the LMM agent, as illustrated in Figure~\ref{fig:prompt_template}. 

For gaining general insights into low-resource harmful meme detection, $X_\text{Reflect}$ is designed as Figure~\ref{fig:X_Reflect}. To restrict the number of insights, we set the capacity of the insight set to 10 by default. Once the insight set is full, the LLM agent is prohibited from producing the \textit{ADD} action.
\begin{figure}
    \centering
    \includegraphics[width=1\linewidth]{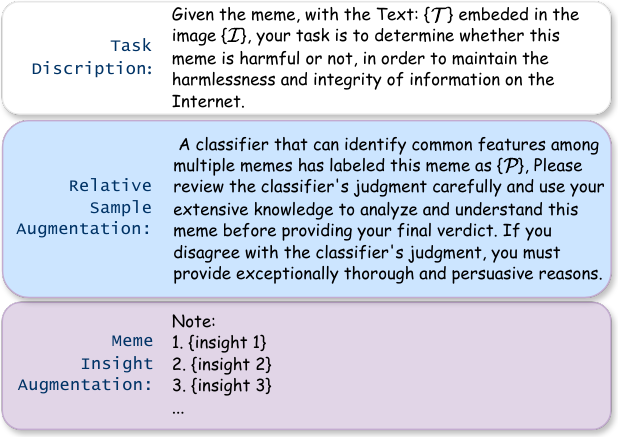}
    \caption{Prompt template for final judgment of \textsc{LoReHM}. }
    \label{fig:prompt_template}
\end{figure}
\begin{figure}
    \centering
    \includegraphics[width=1\linewidth]{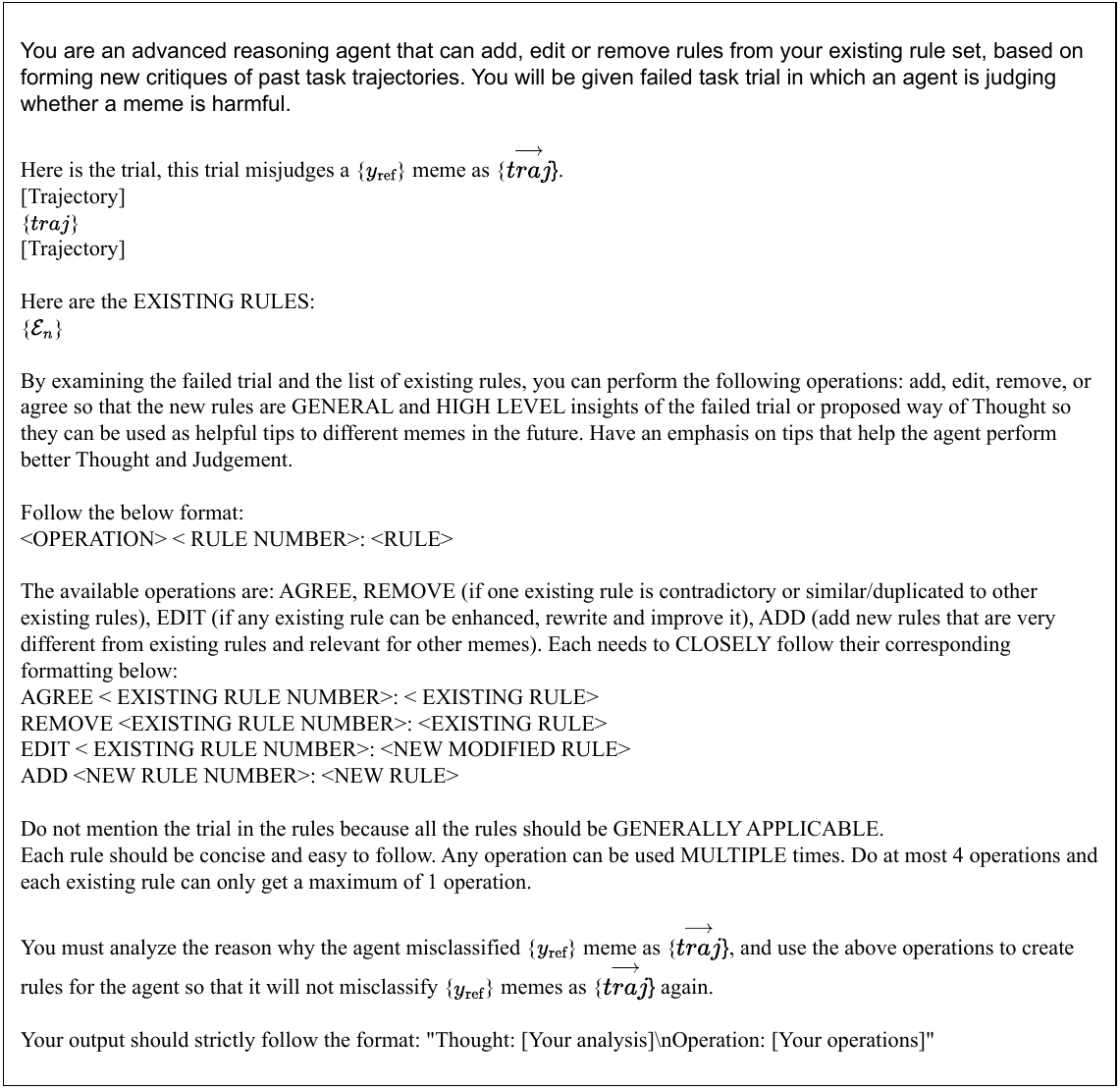}
    \caption{$X_\text{Reflect}$.}
    \label{fig:X_Reflect}
\end{figure}

All experiments were conducted using OpenAI API and four NVIDIA A40 48GiB GPUs.
Evaluation of HarM's test set using \textsc{LoReHM} based on GPT-4o takes approximately 40 minutes, while FHM requires 1 hour and MAMI requires 2 hours. When employing \textsc{LoReHM} based on LLaVA-34B, evaluation times extend to 2 hours for HarM, 3 hours for FHM, and 6 hours for MAMI, approximately.

Due to privacy and ongoing research considerations, the code used in this study is not included in the submission. However, we commit to making the code publicly available upon the acceptance of this paper.

\section{\textsc{LoReHM} Algorithm}
We provide the algorithms for the gradient-free agentic paradigm of our approach, as depicted in Algorithms~\ref{algoritm1}-\ref{algoritm4}.

\begin{algorithm}
\caption{\textsc{LoReHM} - Initialization}
\begin{algorithmic}[0]
\STATE \textbf{Initialize:}
\STATE Ratio factors for visual and textual embeddings $\alpha$, $\beta$;
\STATE Visual Encoder $\operatorname{VE}(\cdot)$;
\STATE Textual Encoder $\operatorname{TE}(\cdot)$;
\STATE Embedding $Emb \leftarrow \emptyset$;
\STATE The visual information $\mathcal{I}$ of a meme $M$;
\STATE The textual information $\mathcal{T}$ of a meme $M$.

\STATE $Emb_{\mathcal{I}} \leftarrow \operatorname{VE}(\mathcal{I})$
\STATE $Emb_{\mathcal{T}} \leftarrow \operatorname{TE}(\mathcal{T})$
\STATE $Emb \leftarrow \alpha \cdot Emb_{\mathcal{I}} + \beta \cdot Emb_{\mathcal{T}}$
\STATE \textbf{return} $Emb$
\end{algorithmic}
\label{algoritm1}
\end{algorithm}

\begin{algorithm}
\caption{\textsc{LoReHM} - Relative Sample Augmentation}
\begin{algorithmic}[0]
\STATE \textbf{Initialize:}
\STATE Feature Representation from Algorithm~\ref{algoritm1}:
\STATE reference meme's embedding information $Emb_\text{ref}$,
\STATE target meme's embedding information $Emb_\text{test}$;

\STATE Set of reference memes $S_{\text{ref}}$;
\STATE Preliminary prediction $\mathcal{P}$;
\STATE Set of Top $K$ similar memes $\mathcal{H} \leftarrow \emptyset$;
\STATE Harmful counter $harmful\_count \leftarrow 0$;
\STATE Similarity scores of two memes $d$;
\STATE Set of Similarity scores $D \leftarrow \emptyset$;
\STATE Function capable of selecting the top $K$ samples in the target set $Top_K (\cdot)$.

\FOR{each $M_{\text{ref}} \in S_{\text{ref}}$}
    \STATE $d \leftarrow \text{cosine}(Emb_{\text{ref}}, Emb_{\text{test}})$
    \STATE $D \leftarrow D \cup d$
\ENDFOR
\FOR{each $M_{\text{ref}} \in S_{\text{ref}}$}
    \IF{$d_\text{ref\_test} \in Top_K(D)$}
        \STATE $\mathcal{H} \leftarrow \mathcal{H} \cup M_{\text{ref}}$
    \ENDIF
\ENDFOR

\FOR{each meme $H_i \in \mathcal{H}$}
    \IF{$H_i$ is harmful}
        \STATE $harmful\_count \leftarrow harmful\_count + 1$
    \ENDIF
\ENDFOR

\IF{$harmful\_count > \frac{K}{2}$}
    \STATE $\mathcal{P} \leftarrow \text{harmful}$
\ELSE
    \STATE $\mathcal{P} \leftarrow \text{harmless}$
\ENDIF

\STATE \textbf{return} $\mathcal{P}$
\end{algorithmic}
\label{algoritm2}
\end{algorithm}

\begin{algorithm}
\caption{\textsc{LoReHM} - Meme Insight Augmentation}
\begin{algorithmic}[0]
\STATE \textbf{Initialize:}
\STATE Reference set $S_{\text{ref}}$;
\STATE Empty insight set $\mathcal{E}_0 \leftarrow \emptyset$;
\STATE Self-reflection set $R_{\text{set}} \leftarrow \emptyset$;
\STATE Trajectory $traj \leftarrow \emptyset$;
\STATE Large Multimodal Model $\operatorname{LMM}$;
\STATE Chain-of-thought prompt $X_{\text{CoT}}$;
\STATE Reflection prompt $X_{\text{Reflect}}$;
\STATE Reference meme's ground truth label $y_{\text{ref}}$;
\STATE Operations produced in insight extraction $O$.

\STATE \textbf{Experience Gathering:}
\FOR{each meme $M_{\text{ref}} \in S_{\text{ref}}$}
    \STATE $traj \leftarrow \operatorname{LMM}(X_{\text{CoT}}, \mathcal{I}_{\text{ref}}, \mathcal{T}_{\text{ref}})$
    \IF{$traj \neq y_{\text{ref}}$}
        \STATE $R_{\text{set}} \leftarrow R_{\text{set}} \cup \{traj\}$
    \ENDIF
\ENDFOR

\STATE \textbf{Insight Extraction:}
\FOR{each $traj_i \in R_{\text{set}}$}
    \STATE $O_i \leftarrow \operatorname{LMM}(X_{\text{Reflect}}, traj_i, \mathcal{E}_{i-1})$
    \STATE $\mathcal{E}_{i} \leftarrow O_i (\mathcal{E}_{i-1})$
\ENDFOR

\STATE \textbf{return} $\mathcal{E}_n$
\end{algorithmic}
\label{algoritm3}
\end{algorithm}

\begin{algorithm}
\caption{\textsc{LoReHM} - Inference}
\begin{algorithmic}[0]
\STATE \textbf{Initialize:}
\STATE Preliminary prediction $\mathcal{P}$ from Algorithm~\ref{algoritm2};
\STATE Insight set $\mathcal{E}_n$ from Algorithm~\ref{algoritm3};
\STATE The visual information $\mathcal{I}_\text{test}$ of a meme $M_\text{test}$;
\STATE The text information $\mathcal{T}_\text{test}$ of a meme $M_\text{test}$;
\STATE Large Multimodal Model $\operatorname{LMM}$;
\STATE Chain-of-thought prompt $X_{\text{CoT}}$;
\STATE Final prediction $p_\text{final}$.

\STATE $p_\text{final} \leftarrow \operatorname{LMM}(X_{\text{CoT}}, \mathcal{I}_\text{test}, \mathcal{T}_\text{test}, \mathcal{P}, \mathcal{E}_n)$

\STATE \textbf{return} $p_\text{final}$
\end{algorithmic}
\label{algoritm4}
\end{algorithm}

\section{Discussion about LMMs}
In this section, we discuss potential concerns on LMMs in the following three aspects: 1) Reproducibility: Since our proposed agency-driven approach uses not only the closed-source GPT-4o but also the open-source LLaVA-34B, the results are definitely reproducible with open-source codes. In order to make sure the LMM could generate the same contents for the same instance, we utilize the specific version ``lava-v1.6-34b'' of LLaVA-34B and ``gpt-4o-2024-05-1'' of GPT-4o, and further set the parameter temperature as 0 without any sampling mechanism, that is, the greedy decoding was adopted to ensure the deterministic results for the content generation with the same prompt. 2) Test Set Leakage: The test set leakage issue does not exist in the open-source LLM~\cite{liu2023visual} as the paper has clearly described the instruction-tuning data used for training, which does not include any data used in our experiments. However, we cannot fully guarantee the exclusion of potential data leakage with GPT-4o, as its internal workings remain opaque. Nevertheless, as evidenced by the results in Table~\ref{tab:ablation}, where the LLaVA-34B or GPT-4o was directly deployed to test on the three standard datasets, the absence of significant test set leakage is implied. This is inferred from the fact that direct zero-shot application of the LMMs did not yield disproportionately high performance, which would be expected if the models were benefiting from test set leakage. Moreover, we can consistently observe enhanced performance in the variants of our proposed framework based on both LLaVA-34B and GPT-4o. This suggests that such improvement is basically attributed to our designed agentic paradigm rather than test set leakage. 3) Generalizability: We believe our \textsc{LoReHM} paradigm is a general technique that works with emerging stronger LMMs, because our approach works not only on GPT-4o, but also well on the open-source LLaVA-34B, which is not an OpenAI system.

\section{Discussion about RSA}
For the proposal of the Relative Sample Augmentation mechanism, we chose not to use the retrieved memes as direct input to the LMM agent. This decision was made to explicitly augment the LMM agent with the label information of the retrieved memes. Simply inputting the retrieved memes back into the LMM agent would merely re-utilize its internal knowledge without incorporating external label information. Additionally, using retrieved memes along with labels as explicit signals is not more effective and somewhat overlaps with the concept of Meme Insight Augmentation. Instead, using a voting mechanism to integrate the label information of retrieved memes as auxiliary signals for outward analysis complements the agent's knowledge-revising strategy through inward analysis. The macro-averaged F1 scores of the voting mechanism are approximately 79\%, 59\%, and 70\% on the HarM, FHM, and MAMI test sets, respectively. Although the voting mechanism itself is not entirely robust, it can still provide effective label information for the LMM agent as a prior preference. From both theoretical and practical perspectives, our current design is the most reasonable and effective approach.

\section{Discussion about MIA}
In our design of Meme Insight Augmentation, the knowledge-revising behavior of human beings is non-trivial when applied to harmful meme detection. Unlike many previous agent tasks that operate within environments providing real feedback, harmful meme detection lacks such an environment to supply the agent with authentic responses for detecting the underlying intent~\cite{lin2021boosting} of the attack. Additionally, because harmful meme detection is fundamentally a binary classification task, informing the agent of an unsuccessful attempt essentially reveals the correct answer. These factors make the direct application of methods like Reflexion~\cite{shinn2024reflexion}, which relies on environmental feedback to determine the success of an attempt and decide whether to reflect and retry based on insights gained from previous attempts, inapplicable. Similarly, ExpeL~\cite{zhao2024expel}, a method requiring multiple passes on the same data to obtain both successful and failed trajectories, cannot be directly utilized. Additionally, during our detailed examination of data samples, we discovered a few instances of contentious annotations. More specifically, in the rare instances of incorrect or disputable annotations, even though we engage in retrieved memes with the incorrect golden label as part of the voting mechanism in the Relative Sample Augmentation phase, the well-generalized insights extracted from the knowledge-revising process also contributes to a better understanding of our approach’s robustness, particularly when dealing with inaccurately annotated data. Therefore, our proposed MIA component is specially devised to couple with the RSA component, which is complementary to each other.


\begin{figure}[t]
    \centering
    \includegraphics[width=1\linewidth]{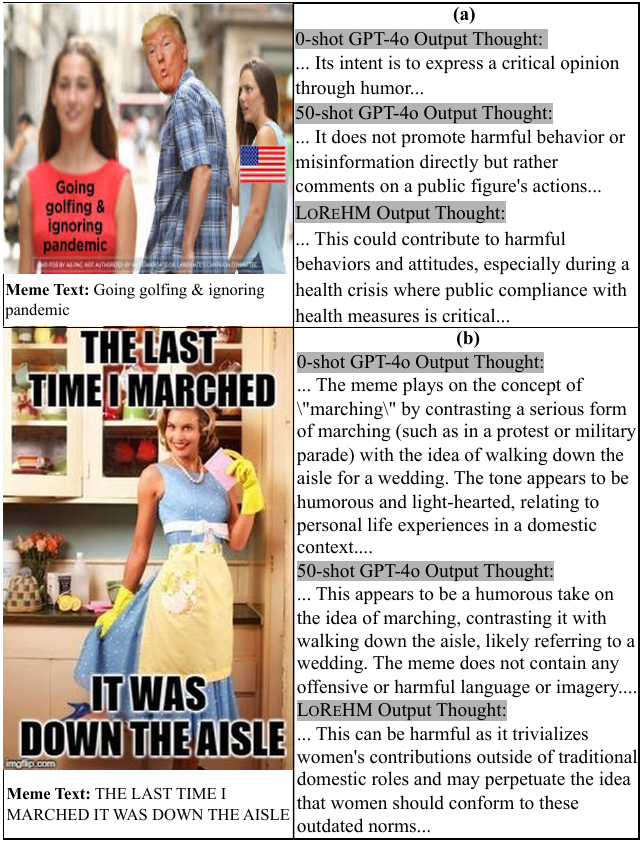}
    \caption{Examples of correctly predicted harmful memes in (a) HarM and (b) MAMI datasets.}
    \label{fig:shots_case}
\end{figure}

\section{More Examples of Case Study}
\label{sec:more_cases}
\subsection{Case Study of Different Labeled Data Sizes}
We provide a case study to compare the different output thoughts from the 0-shot GPT-4o, 50-shot GPT-4o and \textsc{LoReHM}, as shown in Figure~\ref{fig:shots_case}.

From the output thought in natural text, we observe that: 1) Our proposed \textsc{LoReHM} could offer a more profound analysis compared to both the 0-shot and 50-shot GPT-4o methods. For example, in Figure~\ref{fig:shots_case}(a), the 0-shot GPT-4o might deem the meme harmless, failing to grasp its satirical nature, while the 50-shot GPT-4o, though more critical, might not fully consider the meme's broader implications. Our proposed \textsc{LoReHM} identifies the meme's potential to trivialize the pandemic's severity with the phrase ``Going golfing \& ignoring pandemic'' leading us to classify the meme as harmful, as it could undermine the importance of public health measures during a critical time. What's more, in Figure~\ref{fig:shots_case}(b), the 0-shot GPT-4o might overlook the meme's reinforcement of traditional gender roles, while the 50-shot GPT-4o, influenced by limited examples, might not fully appreciate the cultural implications. \textsc{LoReHM}, however, scrutinizes the meme's depiction of a woman in a 1950s-style housewife outfit with the text ``THE LAST TIME I MARCHED IT WAS DOWN THE AISLE'' concluding that it subtly perpetuates outdated gender norms, which can be harmful in the context of modern feminist movements. 2) Our method exhibits notable efficiency within the constraints of low-resource scenarios~\cite{lin2022detect}. Compared to the 50-shot GPT-4o, which relies on a limited set of examples to improve its judgments, our method shows superior performance with the same resource constraints. By effectively leveraging few-shot learning and relevant sample retrieval, our method can discern harmful content with greater accuracy. This capability is crucial when the availability of training examples is scarce, as our method requires fewer examples to achieve a higher level of performance, thus outperforming the 50-shot GPT-4o in equal-resource conditions.

\begin{figure}
    \centering
    \includegraphics[width=1\linewidth]{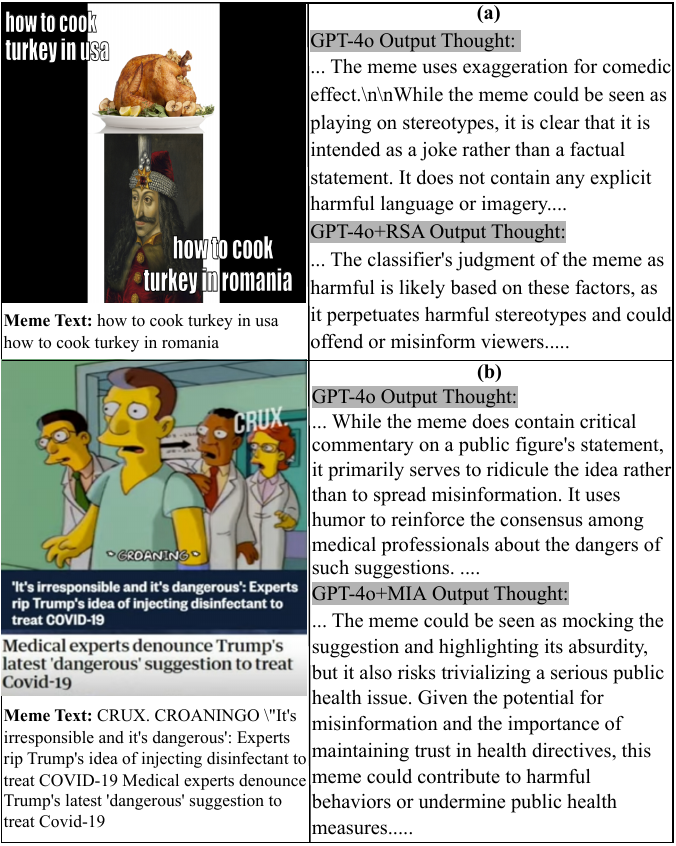}
    \caption{Examples of correctly predicted harmful memes in (a) FHM and (b) HarM datasets.}
    \label{fig:lorehm_split}
\end{figure}

\subsection{Case Study of RSA \& MIA}
We provide a case study of the LMM agent's output thoughts, to investigate the effect of the RSA and MIA strategies on the correctly predicted harmful meme samples by \textsc{LoReHM}, as illustrated in Figure~\ref{fig:lorehm_split}.

From the output thought in natural text, we observe that: 1) The Relative Sample Augmentation (RSA) mechanism enhances the LMM agent by incorporating label information from retrieved memes, providing the LMM agent with additional context and insights. For example, in Figure~\ref{fig:lorehm_split}(a), the GPT-4o output considered it harmless, interpreting it as a joke. However, with RSA, the LMM agent identified it as perpetuating harmful stereotypes, recognizing the potential to offend or misinform viewers. This shows how RSA integrates external labels as auxiliary signals, complementing the agent's internal knowledge-revising strategy and creating a balanced mechanism that leverages both internal and external information sources. By focusing on the labels of retrieved memes rather than directly inputting the memes themselves, RSA ensures that the augmentation process adds unique value to the LMM agent’s capabilities. 2) The Meme Insight Augmentation (MIA) mechanism offers significant benefits by enabling the LMM agent to revise its knowledge effectively. For instance, in Figure~\ref{fig:lorehm_split}(b), the GPT-4o output thought it was a humorous critique and not harmful. However, MIA identified the risk of trivializing a serious public health issue, noting the potential for misinformation that needs to be fact-checked~\cite{lin2022amif} and its impact on public health measures. In emerging events on social media, MIA could promote a knowledge-revising behavior akin to human reasoning, allowing the LMM agent to better interpret memes and distinguish between harmful and harmless content.


\begin{figure}[t]
    \centering
    \includegraphics[width=1\linewidth]{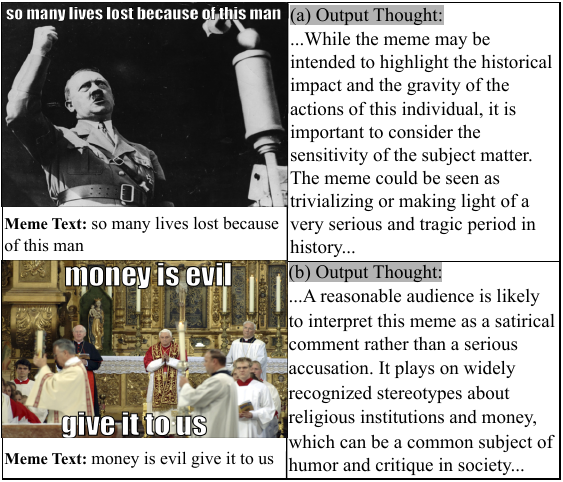}
    \caption{Examples of wrongly predicted memes by our proposed framework with the ground truth (a) harmless and (b) harmful.}
    \label{fig:error_analysis}
\end{figure}

\section{Error Analysis}
\label{sec:error_analysis}
To better understand the behavior of our framework and facilitate future studies, we conduct an error analysis on the wrongly predicted memes by our proposed framework. Figure~\ref{fig:error_analysis} shows two examples of memes wrongly classified by \textsc{LoReHM}. For the harmless meme in Figure~\ref{fig:error_analysis}(a), featuring a historical figure with the text ``so many lives lost because of this man'', \textsc{LoReHM} incorrectly categorized it as harmful. The original thought provided by \textsc{LoReHM} was that the meme could be seen as trivializing a serious and tragic period in history. The misjudgment in this case likely arose from \textsc{LoReHM}'s failure to effectively associate the image with the accompanying text, leading to an overemphasis on the historical figure's negative connotations without considering the full context provided by the meme's message.
In contrast, the harmful meme in Figure~\ref{fig:error_analysis}(b), which depicted several clergymen with the text ``money is evil give it to us'' was deemed harmless by \textsc{LoReHM}.  This misjudgment indicates an inherent bias in \textsc{LoReHM} towards religious-themed memes. Contrary to being overly sensitive, \textsc{LoReHM}'s bias led to an underestimation of the potential negative impact of such content. Additionally, we also investigated more errors resulting from the impact of visual artifacts like image quality, occlusion, obscurity, etc. We found that low-quality images lead to the wrong recognition of superficial patterns and ignoration of the occlusion.

\section{Related Work about LLMs and LMMs}
Recently, LLMs have demonstrated exceptional versatility across various tasks. Significant advancements by leading tech companies have resulted in highly proficient, though often proprietary, LLMs~\citep{brown2020language, OpenAI2023GPT4TR, chowdhery2022palm, team2023gemini}. Meanwhile, the NLP community has seen the rise of open-source LLMs, with publicly shared model weights~\citep{black2022gpt, zeng2022glm, touvron2023llama, touvron2023llama2, luo2023wizardcoder}. More recently, LLMs have also been developed to adapt to processing both textual and visual data, marking a significant advancement. Recent research has focused on constructing versatile multimodal datasets~\citep{yang2023dawn} from platforms like GPT-4 and GPT-4V~\citep{OpenAI2023GPT4TR}, fine-tuning open-source LMMs, \eg, LLaVA~\citep{liu2023visual} and other innovative projects~\citep{bai2023qwen, Dai2023InstructBLIPTG, wang2023cogvlm}. GPT-4o, the latest version of the GPT family, has been released lately. These LMMs have shown excellent emergent abilities in multimodal tasks. In this work, we utilize the representative LMMs, \ie, LLaVA-34B and GPT-4o, as the backbone variants of the LMM agent, to explore the low-resource issue~\cite{lin2023zero} in harmful meme detection with limited labeled meme examples.


\section{Examples of Insights}
In this section, we exemplify the insights drawn from the MIA strategy with 50-shot meme examples.

\paragraph{HarM by \textsc{LoReHM}(GPT-4o)}
\begin{itemize}
    \item Evaluate if the humor in the meme is appropriate and does not trivialize serious issues, create unnecessary panic, target individuals or groups in a derogatory manner, or undermine critical health measures and policies, especially during ongoing crises like pandemics.
    \item Consider the broader context and potential negative impact of the meme, especially in sensitive situations such as health crises or political discourse, to determine if it could cause fear, anxiety, spread misinformation, trivialize serious issues, or undermine important public health measures.
    \item Identify if the meme mocks, trivializes, or undermines public health measures and policies, as this can contribute to harmful behaviors, spread misinformation, and undermine public trust in health directives and governance.
    \item Ensure the meme does not encourage or normalize behavior that could be harmful in real life, particularly in the context of public health and safety, such as trivializing quarantine, policy effectiveness, or other critical health measures.
    \item Assess if the meme perpetuates misinformation or undermines public health measures, which can contribute to harmful outcomes.
    \item Ensure the meme does not inadvertently normalize, trivialize, or encourage harmful behaviors by presenting them humorously, especially in the context of ongoing public health issues and governance policies.
    \item Recognize that memes leveraging fear-based tactics, even for promoting good behavior, can be harmful and should be scrutinized closely.
    \item Assess the potential for the meme to incite derogatory or harmful sentiments towards individuals or groups, which can contribute to a negative, hostile, or unproductive environment.
    \item Assess if the meme trivializes or mocks behaviors that were critical during crises (\eg, panic buying, hoarding) in a way that could diminish the perceived seriousness of the situation or encourage similar behavior in the future.
    \item Determine if the meme's absurdity or exaggeration clearly indicates humor, reducing the likelihood of it being taken seriously and causing harm.
\end{itemize}

\paragraph{FHM by \textsc{LoReHM}(GPT-4o)}
\begin{itemize}
    \item Evaluate the context and possible interpretations of both the image and text together to determine if the meme could be harmful.
    \item Use a balanced approach, considering the potential for harm while recognizing the context, intent, and humor of the meme. Clearly differentiate between harmful content and satire or light-hearted humor to avoid unnecessary censorship.
    \item Assess the potential for a meme to be perceived as offensive by considering the broader context and audience, but avoid overemphasizing potential offense if the context and humor suggest a harmless or light-hearted intent.
    \item Consider the use of exaggeration and satire in memes, recognizing that these elements are often used for humor. Clearly differentiate between satirical humor and content with genuine harmful intent.
    \item Always consider the potential for text in a meme to be interpreted in a harmful or inappropriate manner. However, distinguish between light-hearted humor and genuinely harmful or disrespectful content.
    \item Consider the broader social, cultural, and racial implications of the meme's content to assess its potential harm.
    \item Differentiate between controversial or politically charged content and genuinely harmful content to prevent unnecessary censorship of opinion-based memes.
    \item When evaluating memes, always consider if the content perpetuates racial, cultural, or ethnic stereotypes, as this can be harmful even if the intent seems humorous.
    \item Explicitly differentiate between anthropomorphism and harmful stereotypes about people or specific groups, recognizing that portraying animals with human traits for humor is generally acceptable if it does not convey a negative or harmful message.
    \item When evaluating memes, consider the historical and cultural significance of the references used, ensuring that sensitive topics are not trivialized or used inappropriately.
\end{itemize}

\paragraph{MAMI by \textsc{LoReHM}(GPT-4o)}
\begin{itemize}
    \item Recognize and account for humor or satire, ensuring they are not mistaken for serious harmful content. Consider the context, intent, and whether the humor directly trivializes serious subjects such as violence, sexual behavior, or perpetuates harmful stereotypes.
    \item When evaluating humor, assess whether it perpetuates or trivializes harmful stereotypes, biases, serious life-threatening situations, or explicit content inappropriately. Consider the overall impact on the targeted group, while balancing this with the intent and context of the humor.
    \item When identifying harmful content, evaluate the context, intent behind the language used, and the cultural nuances of humor, especially when dealing with sensitive topics.
    \item Consider the potential for reinforcing harmful stereotypes, real-world harm, or incitement that the content may cause, but also recognize the difference between satire and genuinely harmful content.
    \item Avoid overgeneralizing potentially divisive content as inherently harmful without further examination.
    \item Consider if the meme's humor is based on sensitive or protected characteristics, and the potential negative impact it could have on those groups, especially if it reinforces harmful stereotypes or biases.
    \item Evaluate whether the meme contains humor that could desensitize viewers to serious issues or dangerous situations, and consider the context and intent to determine if it amplifies potential harm.
    \item Evaluate if the humor or satire presents a conflicting message that diminishes the original intent of empowering or positive content.
    \item Distinguish between explicit content meant to entertain a mature audience and content genuinely intended to harm or offend, ensuring cultural and contextual humor is appropriately considered.
    \item Consider whether the combination of humor and serious messages dilutes the impact of the positive message or reinforces harmful attitudes.
\end{itemize}

\paragraph{HarM by \textsc{LoReHM}(LLaVA-34B)}
\begin{itemize}
    \item Assess the overall impact of the meme, including whether it is likely to be perceived as light-hearted or offensive, especially in sensitive contexts such as health crises, disasters, social issues, or activism. Consider whether it trivializes serious situations, promotes irresponsible or dangerous behavior, spreads misinformation, or is insensitive to those affected, even if the intent is humorous.
    \item Consider the context and intent of the meme, especially if it is meant to be humorous or relatable, but also weigh the potential for it to be perceived as insensitive, harmful, or misleading in serious situations.
    \item Evaluate whether the meme directly or indirectly promotes harmful stereotypes, ideas, or personal attacks, trivializes serious issues or individuals’ efforts, spreads misinformation, or could be perceived as insensitive or harmful.
    \item Evaluate whether the meme trivializes serious situations, issues, or the efforts of activists, which could be harmful or offensive to those affected, and consider the potential for it to spread misinformation, encourage irresponsible behavior, or be perceived as insensitive, especially in the context of public health and safety.
    \item Evaluate whether the meme could potentially encourage harmful actions or behaviors, especially in the context of public health and safety. Consider the potential for it to downplay the severity of serious issues, spread misinformation, or be perceived as insensitive, even if the intent is humorous.
    \item Consider the potential for the meme to spread misinformation, harmful stereotypes, or contribute to harmful narratives, even if it is intended as a joke, especially in the context of public health and safety.
    \item Pay special attention to memes that reference sensitive social, political, or cultural issues, as these are more likely to be harmful.
    \item Assess the potential for the meme to be interpreted in a harmful or misleading way by different audiences, considering the broader social, political, and cultural context, and the potential for it to spread misinformation or harmful narratives.
    \item Evaluate the potential for the meme to be misinterpreted by different audiences, leading to harmful consequences, even if the intent is humorous, and consider the broader social and cultural context.
    \item Consider the potential for the meme to be interpreted as promoting or endorsing harmful actions or behaviors, especially in the context of public health and safety.
\end{itemize}

\paragraph{FHM by \textsc{LoReHM}(LLaVA-34B)}
\begin{itemize}
    \item Always consider the potential for double meanings, innuendos, and cultural stereotypes in the text of a meme.
    \item Evaluate the context in which the meme could be shared, its intent, and its potential impact on different audiences.
    \item Be cautious of memes that could be interpreted as promoting or trivializing inappropriate or harmful behavior.
    \item Consider the broader social and cultural implications of the meme's content and text, while also weighing the intent, humor, and factual accuracy behind it.
    \item Be sensitive to cultural stereotypes and avoid memes that could perpetuate or reinforce negative stereotypes about any group.
    \item Differentiate between provocative or controversial statements and genuinely harmful content, ensuring that the meme's potential to cause real harm is the primary consideration.
\end{itemize}

\paragraph{MAMI by \textsc{LoReHM}(LLaVA-34B)}
\begin{itemize}
    \item Always consider if the meme perpetuates stereotypes or objectifies individuals based on gender, race, or other attributes.
    \item Assess if the humor or content used in the meme could be interpreted as demeaning, derogatory, or objectifying towards any group.
    \item Take into account the broader social context, intent, and potential negative interpretations of the meme, distinguishing between harmful content and harmless humor.
    \item Evaluate whether the meme could be perceived as trivializing or making light of serious social issues, even if the intent appears humorous.
    \item Consider the use of satire, exaggeration, and wordplay in the meme and whether it is intended to provoke thought or humor rather than to harm.
\end{itemize}

\section{Examples of Retrieved Memes}
\begin{figure*}
    \centering
    \includegraphics[width=1\linewidth]{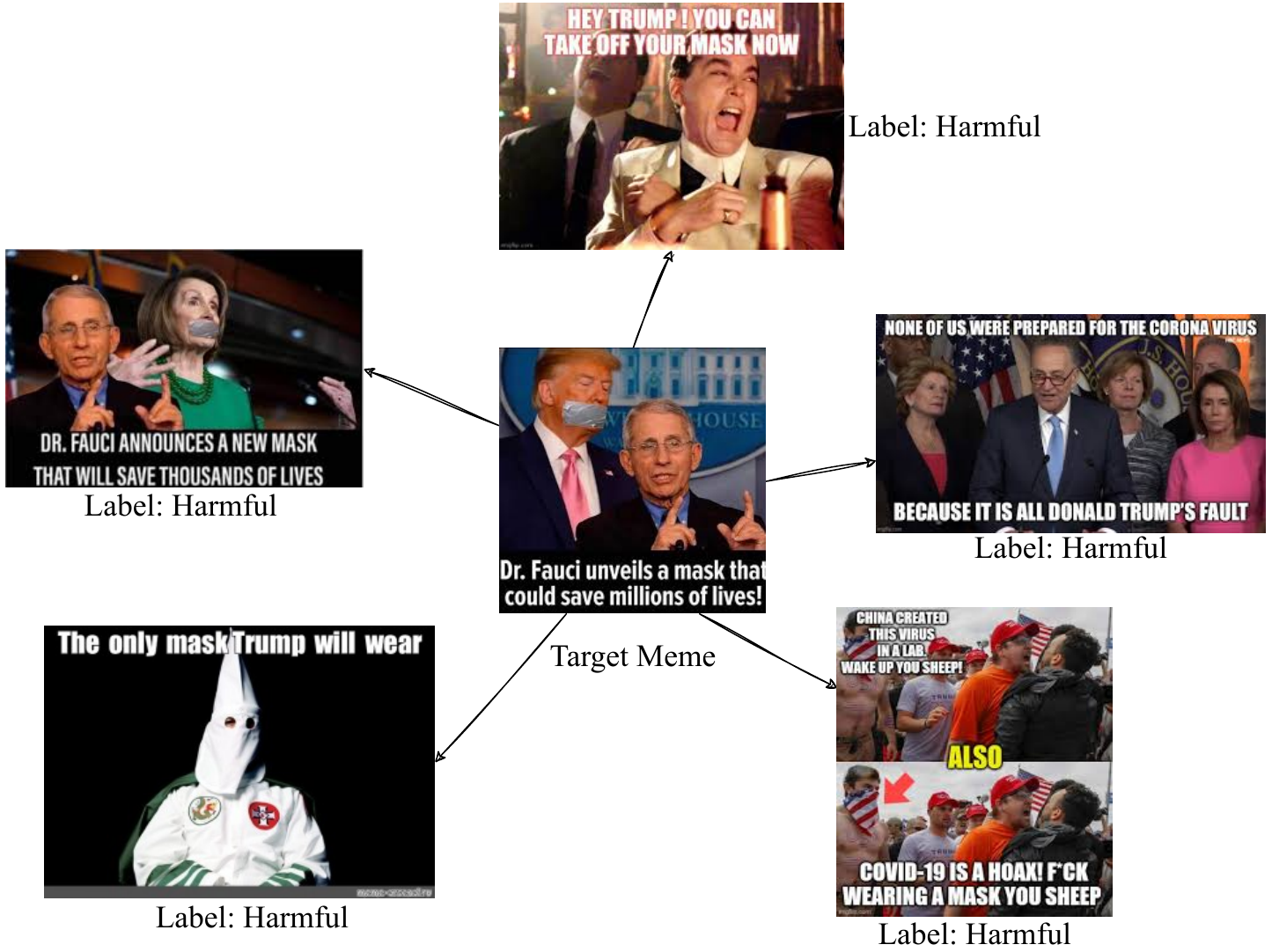}
    \caption{Retrieved top 5 relative memes to the target meme sampled from HarM.}
    \label{fig:retrieval_harm}
    
\end{figure*}

\begin{figure*}
    \centering
    \includegraphics[width=1\linewidth]{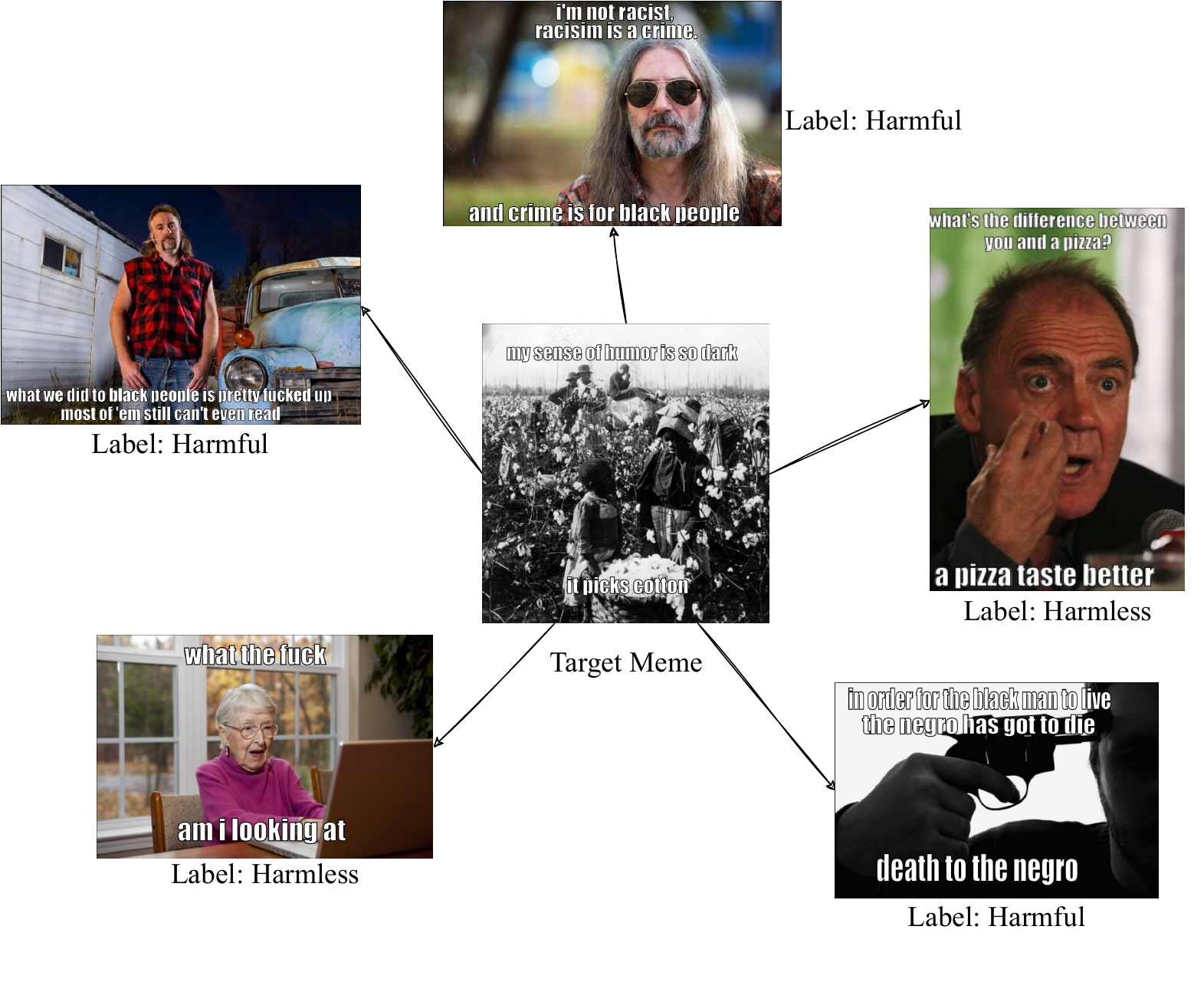}
    \caption{Retrieved top 5 relative memes to the target meme sampled from FHM.}
    \label{fig:retrieval_fhm}
\end{figure*}

\begin{figure*}
    \centering
    \includegraphics[width=1\linewidth]{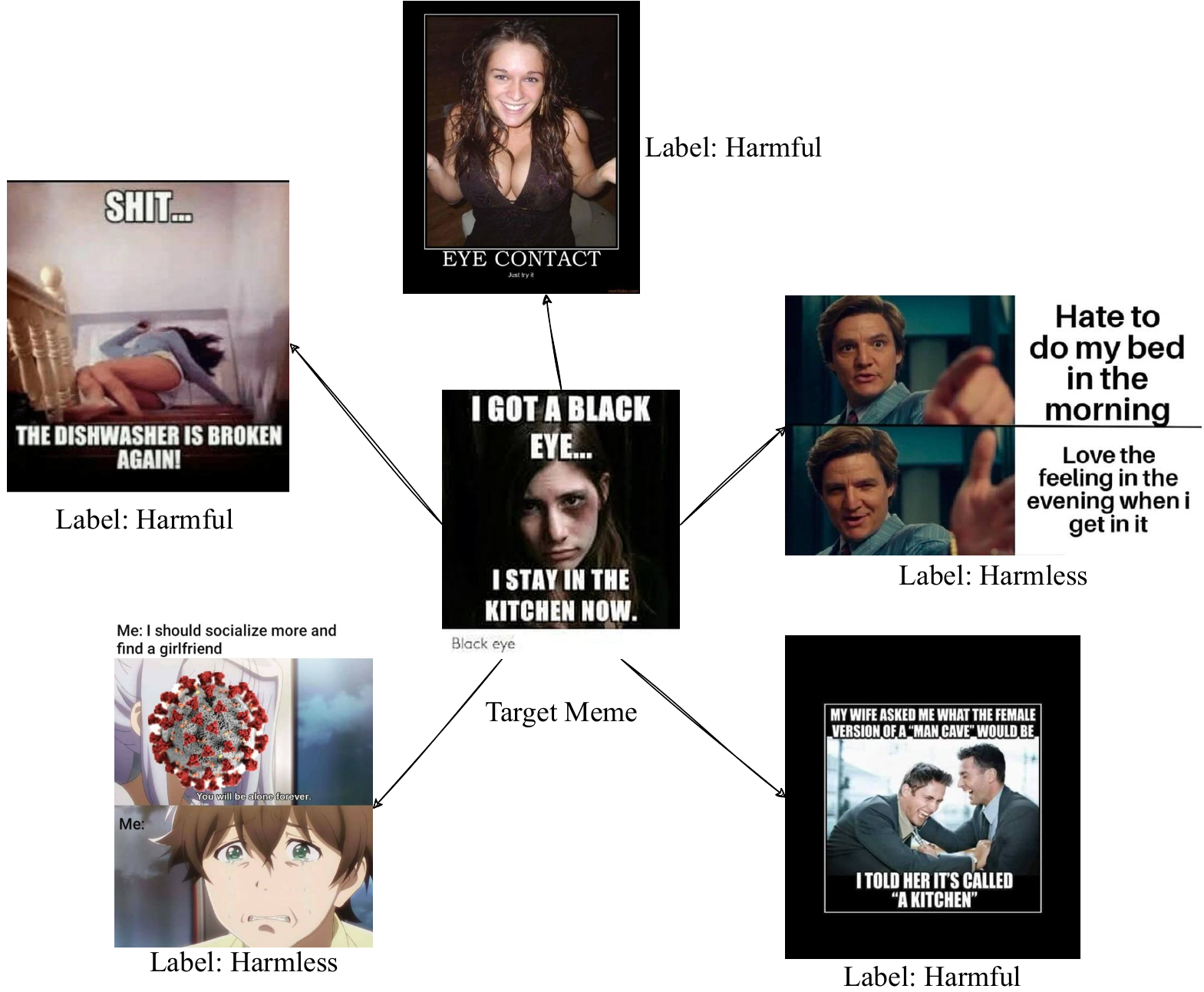}
    \caption{Retrieved top 5 relative memes to the target meme sampled from MAMI.}
    \label{fig:retrieval_mami}
\end{figure*}

Figures~\ref{fig:retrieval_harm}-\ref{fig:retrieval_mami} demonstrate the top 5 relative memes retrieved from the RSA strategy with 50-shot meme examples on HarM, FHM, and MAMI, respectively.

\end{document}